\documentclass[10pt,twocolumn,letterpaper]{article}

\usepackage{iccv}              
\usepackage{makecell} 
\usepackage{multirow}  

\usepackage{bm}
\usepackage{comment}

\usepackage[para]{footmisc}

\definecolor{iccvblue}{rgb}{0.21,0.49,0.74}
\usepackage[pagebackref,breaklinks,colorlinks,allcolors=iccvblue]{hyperref}

\newcommand{\nickname}{RayletDF}

\def\paperID{2858} 
\def\confName{ICCV}
\def\confYear{2025}

\title{\nickname{}: Raylet Distance Fields for Generalizable 3D Surface Reconstruction from Point Clouds or Gaussians}

\author{Shenxing Wei \footnotemark[2] \quad Jinxi Li \footnotemark[2] \quad Yafei Yang \quad Siyuan Zhou \quad Bo Yang \footnotemark[1]\\
vLAR Group, The Hong Kong Polytechnic University\\
{\tt\small \{shenxing.wei, jinxi.li, ya-fei.yang, siyuan.zhou\}@connect.polyu.hk, bo.yang@polyu.edu.hk}
}

\begin{document}

\twocolumn[{%
\renewcommand\twocolumn[1][]{#1}%
    \maketitle
    \begin{center}
        \vspace{-25pt}
        \centering
        \includegraphics[scale=0.73]{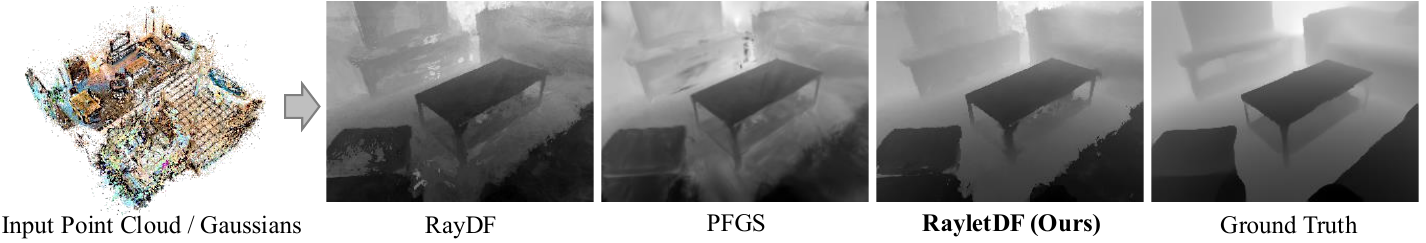}
        \vspace{-20pt}
        \captionof{figure}{We train all methods on the large-scale ARKitScene dataset and then directly test them on the unseen ScanNet++ dataset. Our \nickname{} shows superior generalizability compared to baselines, achieving significantly more accurate 3D scene surface reconstructions.}
        \label{fig:opening}
        \vspace{5pt}
    \end{center}
}]

\renewcommand{\thefootnote}{\fnsymbol{footnote}}
\footnotetext[2]{Equal contribution}
\footnotetext[1]{Corresponding author}

\begin{abstract}
In this paper, we present a generalizable method for 3D surface reconstruction from raw point clouds or pre-estimated 3D Gaussians by 3DGS from RGB images. Unlike existing coordinate-based methods which are often computationally intensive when rendering explicit surfaces, our proposed method, named \textbf{\nickname{}}, introduces a new technique called raylet distance field, which aims to directly predict surface points from query rays. Our pipeline consists of three key modules: a raylet feature extractor, a raylet distance field predictor, and a multi-raylet blender. These components work together to extract fine-grained local geometric features, predict raylet distances, and aggregate multiple predictions to reconstruct precise surface points. We extensively evaluate our method on multiple public real-world datasets, demonstrating superior performance in surface reconstruction from point clouds or 3D Gaussians. Most notably, our method achieves exceptional generalization ability, successfully recovering 3D surfaces in a single-forward pass across unseen datasets in testing. Our code and datasets are available at {\small\url{https://github.com/vLAR-group/RayletDF}}.
\end{abstract}   

\section{Introduction}
\label{sec:intro}
Learning efficient, accurate, and generalizable 3D surface representations is crucial for many applications in mixed reality, embodied AI, and graphics. Given RGB/D images and/or point clouds, a series of 3D representations has been developed to recover 3D geometry, including occupancy fields (OF) \cite{Mescheder2019}, un/signed distance fields (U/SDF) \cite{Chibane2020a,Park2019}, radiance fields (NeRF) \cite{Mildenhall2020}, and vector fields (VF) \cite{Yang2023}. While achieving excellent results in 3D reconstruction, these coordinate-based methods and their variants typically demand dense network evaluations to obtain explicit surfaces, thus being computationally heavy. 

Recently, the point-based method 3D Gaussian Splatting (3DGS) \cite{Kerbl2023} has emerged as an appealing alternative to those coordinate-based methods, thanks to its impressive real-time performance in synthesizing high-fidelity RGB images at high resolutions. However, it still falls short in rendering high-quality depth views, due to its failure in capturing fine-grained surface geometry, though various constraints such as depth priors \cite{Li2024,Wolf2024}, local smoothness or planarity regularization \cite{Wu2024,Guedon2024} can be added to enhance the quality of surfaces recovered. 

On the other hand, a line of recent ray-based methods such as DRDF \cite{Kulkarni2022}, PRIF \cite{Feng2022}, DDFs \cite{Aumentado-Armstrong2022}, Pointersect \cite{Chang2023}, and RayDF \cite{Liu2023a} have demonstrated excellent performance in representing intricate surfaces based on light rays. Nevertheless, due to the limitation of existing ray parametrizations such as Plucker and spherical coordinates, they are often limited to recovering object-level surfaces and require per-scene training, lacking the desired generalizability to infer diverse 3D scene structures in a single forward pass. 

In this paper, we present a generalizable 3D surface representation pipeline to accurately recover 3D geometry. The core module of our pipeline is a new technique called raylet distance field, with inspiration from ray-based methods. 
\begin{figure}[t]\vspace{-0.2cm}
\setlength{\abovecaptionskip}{ 4. pt}
\setlength{\belowcaptionskip}{ -10 pt}
\centering
   \includegraphics[width=0.65\linewidth]{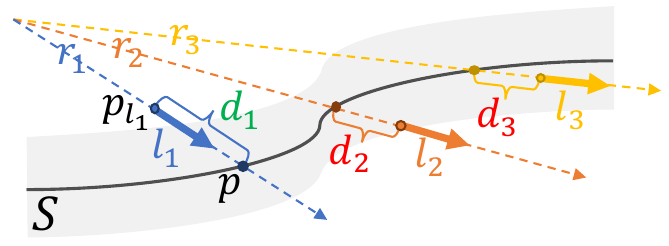}
\caption{An illustration of raylets and raylet distances.}
\label{fig:raylet_rayletDistance}
\vspace{-0.3cm}
\end{figure}

The term \textit{raylet} is defined as a unit segment of a light ray and its starting point is sampled (or located) near the surface of a shape. The \textit{raylet distance} is defined as the signed distance between the surface hit point and the raylet starting point. As illustrated in Figure \ref{fig:raylet_rayletDistance}, given three light rays $\{\bm{r}_1, \bm{r}_2, \bm{r}_3\}$ emitting from  light sources, we sample three raylets $\{\bm{l}_1, \bm{l}_2, \bm{l}_3\}$ near the surface $\bm{S}$, where raylet distance $d_1$ is assigned positive as its surface hit point $\bm{p}$ locates further away from the raylet starting point $\bm{p}_{l_1}$, whereas $\{d_2, d_3\}$ are assigned negative accordingly. Note that, given a single light ray, we can sample numerous raylets on both sides of the surface hit point. Intuitively, such definitions of raylets and raylet distances primarily focus on nuanced local patterns of surfaces, and these local geometric patterns are usually generalizable across diverse shapes. 

With this merit of raylets, we simply formulate the problem of generalizable 3D surface reconstruction into learning raylet distance fields from visual observations. In particular, our pipeline comprises three modules: 1) a \textbf{raylet feature extractor} to extract geometry features from an input 3D scene for a query raylet; 2) a \textbf{raylet distance field} to predict the signed raylet-surface distance value for the input query raylet; and 3) a \textbf{multi-raylet blender} to aggregate multiple predicted raylet distances along every single ray, ultimately recovering the accurate surface point. 

These three modules together allow us to learn generalizable ray-based surface representations from large-scale training data with ground truth depth scans as supervision signals. Once the pipeline is well trained, it can be used to infer high-quality 3D scene surfaces on unseen datasets from arbitrary query viewing angles in a single forward pass. Notably, our pipeline can not only train and evaluate on point clouds, but is also amenable to any scene represented by 3D Gaussians recovered by the popular 3DGS \cite{Kerbl2023} from RGB images. Our pipeline is called \textbf{\nickname{}} and Figure \ref{fig:opening} shows qualitative results on diverse indoor scenes. Our contributions are:
\begin{itemize}[leftmargin=*]
\setlength{\itemsep}{1pt}
\setlength{\parsep}{2pt}
\setlength{\parskip}{1pt}
    \item We propose a generic pipeline for explicit 3D surface reconstruction from either point clouds or 3D Gaussians. 
    \item We introduce a new raylet distance field followed by a blender to learn generalizable surface patterns. 
    \item We show superior accuracy in surface reconstruction across multiple datasets, clearly outperforming baselines, especially in terms of generalizability to new datasets.
    \item We benchmark generalizable surface reconstruction on 3D Gaussians and will release 7770 3D scene Gaussians trained on ScanNet/++, ARKit, and MultiScan datasets.
\end{itemize}

\section{Related Works}
\label{sec:literature}

Classical approaches to recovering 3D surfaces from images mainly include SfM \cite{Ozyesil2017} and SLAM \cite{Cadena2016} systems to obtain sparse point clouds, optionally followed by various global or local smoothness priors to recover continuous surfaces, but the reconstructed shapes usually lack fine details. Early learning methods to model explicit 3D structures mainly include voxel grids \cite{Chan2016,Yang2018,Yang2020}, point clouds \cite{Fan2017}, octree \cite{Tatarchenko2017}, meshes \cite{Kato2017} based pipelines, but the fidelity of these discrete shape representations is often limited by the spatial resolutions and memory footprint. A comprehensive survey of these methods can be found in \cite{Berger2017a,Han2019}.

\textbf{Coordinate-based Methods for Surface Reconstruction}: To avoid the discretization issue of classical explicit 3D representations, a series of implicit representations has been developed to use simple MLPs to recover continuous 3D shapes \cite{Fried2020}. These representations, including OF \cite{Chen2019g,Mescheder2019}, SDF \cite{Park2019}, UDF \cite{Chibane2020a,Wang2022}, VF \cite{Yang2023,Rella2024}, and NeRF \cite{Mildenhall2020,Trevithick2021}, typically take 3D coordinates as input and predict various properties of input coordinates. While demonstrating exceptional accuracy in recovering continuous 3D surfaces and/or rendering 2D views, these methods and their variants \cite{Wang2024,Tewari2021,Xie2022} demand dense 3D point sampling and network evaluations to regress explicit surface points given any query view, thus being computationally heavy. In this paper, our \nickname{} learns efficient raylet fields and does not require dense sampling to regress surface points. 

\textbf{Point-based Methods for Surface Reconstruction}: Point-based methods aim to recover surfaces by rendering disconnected geometry samples, including the simplest point sample \cite{Grossman1998} and circular/ elliptic discs/ ellipsoids/ surfels sample rendering methods \cite{Pfister2000,Ren2003,Zwicker2001}. Recently, a new point-based method 3DGS \cite{Kerbl2023} has been introduced to represent the scene by a set of 3D Gaussians with various properties such as position, covariance, and color. While achieving real-time rendering of 2D views thanks to the tile-based rasterizer, 3DGS falls short in synthesizing highly accurate depth views as it is hard to capture fine-grained surfaces, though advanced techniques \cite{Chen2024,Fan2024,Wu2024,Guedon2024,Huang2024,Chung2024,Xu2025,Li2024,Huang2025,Yu2024a,Xiang2024,Ouyang2024,Yu2024,Song2024,Chen2024a,Zhang2024,Zhang2024a} can mitigate this issue but at the expense of generalization across various scenes. Meanwhile, researchers also introduce fast winding number to balance the efficiency and accuracy \cite{barill2018fast,chen20243d}. These approaches achieve better geometry fidelity and convergence speed but still need per-scene optimization and point normals.
In this paper, our introduced raylet distance field and the blender aim to learn nuanced local surface features, thus being accurate and generalizable.  

\begin{figure*}[t]
\centering
\includegraphics[width=0.99\linewidth]{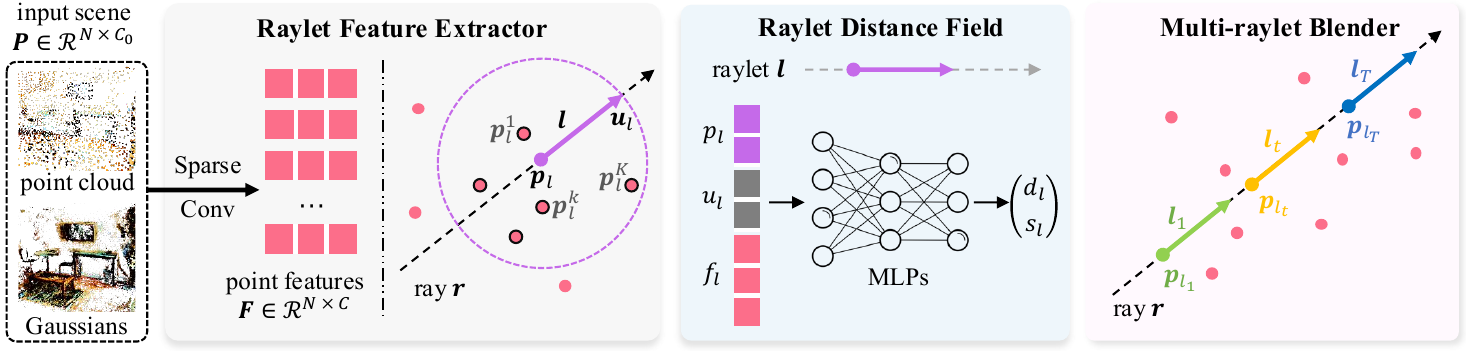}
\caption{Overview of our proposed pipeline. The leftmost block shows the raylet feature extractor module, the middle block shows the raylet distance field module, and the rightmost block shows our multi-raylet blender module.}
\label{fig:network}
\vspace{-0.4cm}
\end{figure*}

\textbf{Ray-based Methods for Surface Reconstruction}: A line of works formulates 3D shapes as ray-based neural functions. Existing ray-based methods simply take individual rays as input and estimate either RGB values such as LFN \cite{Sitzmann2021}, NeuLF \cite{Liu2021a} and others \cite{Feng2021,Ost2022,Suhail2022,Cao2023}, or surface points including PRIF \cite{Feng2022}, DDFs \cite{Aumentado-Armstrong2022}, RayDF \cite{Liu2023a}, and others \cite{Kulkarni2022,Sundt2023,Jin2024,Sundt2024,Yin2024}. Although these methods have shown excellent efficiency and accuracy in rendering surfaces, they can typically represent object-level shapes and lack generality due to the limitation of light ray representations. In this paper, instead of using global light rays to represent surfaces, we use raylets to recover geometry. 

\textbf{Image-based Depth Estimation}: With the advancement of techniques for novel view RGB synthesis, existing methods \cite{Ranftl2022,Bochkovskii2025,Yang2024} for image-based depth estimation can also be used to recover per-view depth scans, followed by depth fusion for 3D surface reconstruction. However, such a pipeline inherently lacks consistent depth alignment and geometry consistency priors. Therefore, the recovered 3D structures are inferior in quality.

\section{\nickname{}}
\label{sec:method}
\subsection{Overview}

Our \nickname{} models 3D surface as a neural network $f$. It takes a scene volumetric data $\bm{P}$ and a query raylet $\bm{l}$ as input, directly predicting the corresponding raylet distance $d_l$, \ie{}, the distance between surface hit point and raylet starting point as shown in Figure \ref{fig:raylet_rayletDistance}. The scene volumetric data can be a sparse point cloud or on-the-shelf 3D Gaussians recovered by the popular 3DGS \cite{Kerbl2023} from RGBs, both denoted by $\bm{P}\in \mathcal{R}^{N\times C_0}$ for simplicity, where $N$ represents the total number of surface points or Gaussian centers and $C_0$ represents the number of per-point features such as point/Gaussian center $xyz$ optionally with other properties like colors or opacity. The query raylet $\bm{l}\in\mathcal{R}^{1\times6}$ is parameterized by its starting point $xyz$ and a unit vector representing ray direction. Formally, it is defined as below:
\begin{equation}
    d_l = f(\bm{P}, \bm{l}) \quad \quad \bm{P}\in \mathcal{R}^{N\times C_0}, \quad \bm{l}\in \mathcal{R}^{1\times6}
\end{equation}

Since our method aims at representing accurate and generalizable 3D surface representations, the pipeline is designed to be trained on a large number of 3D scenes with depth images as supervision signals. Once well-trained, it is expected to be applied to various unseen datasets, directly inferring high-quality depth maps at arbitrary query viewing angles. As shown in Figure \ref{fig:network}, our pipeline has three components clarified in the following Sections \ref{sec:raylet_feat_extract}\&\ref{sec:raylet_field}\&\ref{sec:multiraylet_blender}. 

\subsection{Raylet Feature Extractor}\label{sec:raylet_feat_extract}
This module is designed to learn local features from an input scene volumetric data point $\bm{P}$ for any specific query raylet $\bm{l}$. As shown in the leftmost block of Figure \ref{fig:network}, given an input scene $\bm{P}$, we simply adopt the powerful SparseConv architecture \cite{Graham2018} without any pretraining step as the backbone network, obtaining per-point features $\bm{F} \in \mathcal{R}^{N\times C}$ where the embedding length $C$ is predefined as 32. Implementation details are in Appendix~\ref{app_spconv}. 

Having the input scene $\bm{P}$ and its per-point features $\bm{F}$, we need to extract the local features for a specific input query raylet $\bm{l}$ whose starting point is denoted by $\bm{p}_l$ and ray direction by a unit vector $\bm{u}_l$. In particular, as shown in Figure \ref{fig:network}, for the query raylet $\bm{l}$, we identify the top $K$ nearest points $\{\bm{p}_l^1\cdots \bm{p}_l^k \cdots \bm{p}_l^K\}$ in the input scene $\bm{P}$ with regard to the raylet starting point $\bm{p}_l$. The simple K-nearest neighbors (KNN) algorithm is adopted based on the point-wise Euclidean distances. After that, we gather the information, denoted by $\bm{\hat{f}}_l^k$, for the $k^{th}$ neighboring point as follows:
\begin{equation}\label{eq:feat_neighbor_point}
\small
    \bm{\hat{f}}_l^k = \Big(\boldsymbol{p}_l^k \oplus\frac{(\boldsymbol{p}_l^k - \boldsymbol{p}_l)}{||\boldsymbol{p}_l^k - \boldsymbol{p}_l ||} \oplus ||\boldsymbol{p}_l^k - \boldsymbol{p}_l || \Big)\oplus \bm{f}_l^k
\end{equation}
where $\boldsymbol{p}_l$ and $\boldsymbol{p}_l^k$ are the \textit{xyz} positions of points, $\oplus$ is the concatenation operation, $||\cdot||$ calculates the Euclidean distance between two points, and $\bm{f}_l^k$ is the feature vector extracted from per-point features $\bm{F}$. 

Lastly, we stack all information of the total $K$ neighboring points into a local feature vector $\bm{f}_l$ for the query raylet $\bm{l}$ for simplicity as follows, though other methods such as max/mean and attentional pooling are also applicable.  
\begin{equation}\label{eq:feat_neighbor_point_No}
    \bm{f}_l = \bm{\hat{f}}_l^1 \oplus \cdots 
\bm{\hat{f}}_l^k \cdots \oplus \bm{\hat{f}}_l^K  
\end{equation}

Overall, our key insight of this module is that the features extracted for a query raylet should retain local geometrical patterns near surfaces, allowing the learned raylet distance representations to be generalizable across diverse scenes.

\subsection{Raylet Distance Field}\label{sec:raylet_field}
This module aims to learn accurate raylet surface distance value for any specific query raylet $\bm{l}$. As illustrated in the middle block of Figure \ref{fig:network}, we feed the query raylet, \ie, the starting point $\bm{p}_l$ and direction $\bm{u}_l$, and its feature vector $\bm{f}_l$ (ref to Section \ref{sec:raylet_feat_extract}) into MLPs, directly predicting the corresponding raylet surface distance value $d_l$. 

To improve the generalizability and robustness of  raylet distance predictions, we further predict a confidence score $s_l$ alongside the distance value $d_l$. 
This score is basically designed to weight the prediction and blend multiple raylets sampled along a single ray which will be detailed in Section \ref{sec:multiraylet_blender}. Formally, the raylet distance field is defined as:
\begin{equation}\label{eq:raylet_dis_field}
    (d_l, s_l) = MLPs\big(\bm{p}_l \oplus \bm{u}_l \oplus \bm{f}_l\big)
\end{equation}

For fast computation, the MLPs consist of 8 layers and each layer has 256 hidden neurons, though more sophisticated neural layers can be applied as well. Since the predicted distance value could be positive or negative, there is no non-linear activation function added at the output layer. More details are in Appendix~\ref{app_raylet_mlp}.

Our design of this raylet distance field is particularly simple in concept, but has a clear advantage over the prevailing coordinate-based surface representations such as OF, SDF, and UDF: our method takes raylets as input, without requiring sampling dense coordinates along a single ray, thus being efficient to regress surface points.

\subsection{Multi-raylet Blender}\label{sec:multiraylet_blender}
The ultimate goal of our pipeline is to infer the surface point for any given light ray $\bm{r}$, \ie, the distance $D$ between the query camera center $\bm{p}_{cam}$ and the surface hit point along the ray. Since our raylet distance field learns representations for raylets instead of entire rays, this allows us to sample multiple raylets along each single ray followed by blending multiple estimations, potentially improving the generalizability and robustness of our method.

To this end, as shown in the rightmost block of Figure \ref{fig:network}, we opt for sampling $T$ raylets $\{\bm{l}_1 \cdots \bm{l}_t \cdots \bm{l}_T\}$ near the surface along a single ray $\bm{r}$. These $T$ raylets have the same unit vector of ray direction, but with different starting point positions, denoted by $\{\bm{p}_{l_1} \cdots \bm{p}_{l_t} \cdots \bm{p}_{l_T}\}$. Naturally, we feed these $T$ raylets into our raylet distance field in parallel, obtaining the corresponding distances and scores: $\{(d_{l_1}, s_{l_1})\cdots (d_{l_t}, s_{l_t}) \cdots (d_{l_T}, s_{l_T})\}$. After that, we blend all these predictions and estimate the distance between camera center $\bm{p}_{cam}$ and surface hit point as follows:
\begin{equation}\label{eq:multi_raylet_blender}
    D = \sum_{t=1}^T \hat{s}_{l_t}\Big(|| \bm{p}_{cam} - \bm{p}_{l_t} || + d_{l_T} \Big), \hat{s}_{l_t} = \frac{e^{s_{l_t}}}{\sum_{t=1}^T e^{s_{l_t}} }
\end{equation}
where $\hat{s}_{l_t}$ is a normalized score. $\bm{p}_{cam}$ and $\bm{p}_{l_t}$ are always known during sampling in training and test. 

The prediction $D$ is fully supervised using $\ell_1$ loss, and the ground truth distance values are converted from the available depth images in training set.

\begin{figure}[t]\vspace{-0.2cm}
\setlength{\abovecaptionskip}{ 1. pt}
\setlength{\belowcaptionskip}{ -18 pt}
\centering
   \includegraphics[width=0.9\linewidth]{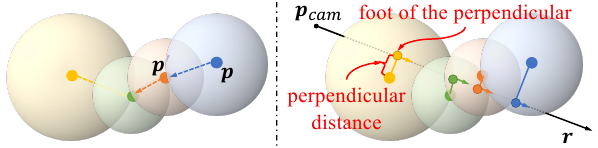}
\caption{Raylet sampling on virtual balls of point clouds.}
\label{fig:raylet_sampling}
\end{figure}

\subsection{Sampling Raylets for Training and Test}\label{sec:sampling_raylets} 
Given a specific 3D scene $\bm{P}$ as input, if it is a raw point cloud, for a specific query ray $\bm{r}$, we sample multiple raylets for both training or test in the following steps:
\begin{itemize}[leftmargin=*]
    \item Step 1: For every point $\bm{p}$ in the input scene point cloud $\bm{P}$, we search its nearest point $\bm{p'}$ and identify the Euclidean distance $|| \bm{p} - \bm{p'}||$ as a radius of the ball centering at point $\bm{p}$. In this way, every point in $\bm{P}$ is assigned a virtual ball. Intuitively, the underlying continuous 3D scene surface, which is unknown and yet to be estimated, should be bounded by the union space of all virtual spheres, as illustrated in the left block of Figure \ref{fig:raylet_sampling}.   
    \item Step 2: as illustrated in the right block of Figure \ref{fig:raylet_sampling}, given the query ray $\bm{r}$ with a camera center $\bm{p}_{cam}$, it shoots through multiple virtual balls. We then project the center point of each intersected ball onto the ray, obtaining a perpendicular distance and the foot of the perpendicular. After that, we select $T$ foot points, belonging to intersected virtual balls with top $T$ shortest perpendicular distances, as the starting points of $T$ raylets. If the number of intersected balls is fewer than $T$, we just keep the actual number of raylets. Note that, if there is no ball intersected, meaning that the ray shoots outside the target 3D surface, the ray is discarded in both training and test. 
\end{itemize} 

If the input 3D scene $\bm{P}$ is a set of 3D Gaussians recovered by 3DGS \cite{Kerbl2023} from RGBs, we follow the technique \cite{Yu2024a,Keselman2022} to calculate an intersection point between the query ray and each Gaussian, and then we select $T$ intersection points, belonging to the Gaussians with top $T$ contributions based on alpha blending, as the starting points of $T$ raylets. 

For efficiency, we follow the tile-based splitting strategy in 3DGS \cite{Kerbl2023} to calculate intersection points on both point clouds and Gaussians. More details are in Appendix~\ref{app_cal_intersect}.

\subsection{Surface Normal Derivation}
Similar to RayDF, our method also has a nice property of deriving a closed-form expression for a normal vector, thanks to the nature of our ray-based formulation. In particular, given an input raylet $\bm{l} = (\bm{p}_l, \bm{u}_l)$ and its estimated raylet distance value $d_l$ from the network, the corresponding normal vector $\bm{n}_l$ is provided in Appendix~\ref{app_normal}. 

This normal vector can be used in an additional loss to regularize surface points, or acts as a threshold to remove outlier predictions, which are left for future exploration.

\begin{table*}[t]\tabcolsep=0.1cm 
\centering
\caption{Quantitative results of all methods for ray surface distance estimation. $^\dagger$ indicates the scale is aligned with ground truth depth.}\vspace{-0.2cm}
  \resizebox{1\textwidth}{!}{
\begin{tabular}{@{}r|c|ccccc|ccccc|ccccc@{}}
    \toprule
     \multicolumn{2}{c|}{\makecell{}} &\multicolumn{5}{c|}{\textit{test on} $\rightarrow$ ARKitScenes}&\multicolumn{5}{c|}{\makecell{\textit{test on} $\rightarrow$ ScanNet/ScanNet++}}&\multicolumn{5}{c}{\makecell{\textit{test on} $\rightarrow$ MultiScan}}\\
   \cmidrule{3-17}
   \multicolumn{2}{c|}{\makecell{}}&ADE $\downarrow$ &RMSE$\downarrow$ &Abs-Rel$\downarrow$&Sq-Rel$\downarrow$&$\delta\uparrow $&ADE$\downarrow$&RMSE$\downarrow$&Abs-Rel$\downarrow$&Sq-Rel$\downarrow$&$\delta\uparrow$&
   ADE$\downarrow$&RMSE$\downarrow$&Abs-Rel$\downarrow$&Sq-Rel$\downarrow$&$\delta\uparrow$ \\
   \midrule
   3DGS~\cite{Kerbl2023}& \multirow{6}{*}{\rotatebox{90}{ \makecell{ \textit{per-scene}\\ train / test}}}
   &0.268&0.379&0.173&0.098&0.707&   0.321&0.461&0.152&0.149&0.761& 0.431&0.532&0.322&0.225&0.402  \\
   GOF~\cite{Yu2024a}&
   &0.318&0.469&0.213&0.159&0.659& 0.320&0.507&0.155&0.214&0.775 &0.571&0.700&0.404&0.373&0.363\\
   PGSR~\cite{Chen2024a}&
   &0.219&0.376&0.168&0.183&0.793& 0.202&0.536&0.099&0.379&0.877 &0.315&0.430&0.285&0.516&0.567\\
   DepthAnythingV2$^\dagger$~\cite{Yang2024}&
   &0.206&0.294&0.144&0.093&0.860& 0.168&0.248& 0.118&0.066&0.887 &0.228&0.303&0.205&0.117&0.775\\
   Depth-Pro$^\dagger$~\cite{Bochkovskii2025}&
   &0.294&0.403&0.202&0.242&0.750& 0.220&0.319&0.120&0.106&0.861 &0.280&0.363&0.234&0.157&0.659 \\
   \midrule
   MVSGaussian~\cite{Liu2024}&\multirow{5}{*}{\rotatebox{90}{ \makecell{\textit{train on} \\ ARKitScenes}}}
   &0.584&0.755&0.392&0.416&0.425 &0.592&0.797&0.395&0.376&0.538 &0.543&0.676&0.444&0.434&0.363\\
   Pointersect~\cite{Chang2023}&
   &0.286&0.397&0.209&0.135&0.749& 0.366&0.513&0.190&0.151&0.733& 0.266&0.351&0.217&0.117&0.671\\
   RayDF~\cite{Liu2023a}&
   &0.183&0.303&0.139&0.088&0.846& 0.227&0.363&0.131&0.123&0.838& 0.326&0.425&0.287&0.202&0.583\\
   PFGS~\cite{Wang2024a}&
   &0.264&0.386&0.171&0.098&0.719 &0.303&0.455&0.145&0.145&0.779 &0.536&0.653&0.442&0.375&0.316\\
   \textbf{\nickname{} (Ours)} & 
   &\textbf{0.115}&\textbf{0.218}&\textbf{0.082}&\textbf{0.047}&\textbf{0.928} &\textbf{0.175}&\textbf{0.320}&\textbf{0.085}&\textbf{0.098}&\textbf{0.894} &\textbf{0.216}&\textbf{0.311}&\textbf{0.182}&\textbf{0.107}&\textbf{0.739}
   \\
   \midrule
    MVSGaussian~\cite{Liu2024}&\multirow{5}{*}{\rotatebox{90}{\makecell{\textit{train on} \\ ScanNet / ++}}}
    &0.600&0.774&0.404&0.448&0.413 &0.591&0.787&0.295&0.367&0.528&0.548&0.681&0.448&0.445&0.363\\
   Pointersect~\cite{Chang2023}&
   &0.328&0.462&0.196&0.142&0.665 &0.433&0.604&0.198&0.199&0.658& 0.404&0.504&0.296&0.198&0.449\\
   RayDF~\cite{Liu2023a}&
   &0.587&0.704&0.348&0.302&0.317 &0.202&0.337&0.114&0.818&0.871 &0.604&0.690&0.447&0.467&0.233\\
   PFGS~\cite{Wang2024a}&
   &0.274&0.399&0.177&0.104&0.704 &0.286&0.433&0.137&0.142&0.798 &0.534&0.669&0.448&2.449&0.327\\
   \textbf{\nickname{} (Ours)}  & 
   &\textbf{0.132}&\textbf{0.241}&\textbf{0.094}&\textbf{0.053}&\textbf{0.906} &\textbf{0.145}&\textbf{0.276}&\textbf{0.072}&\textbf{0.079}&\textbf{0.922} &\textbf{0.259}&\textbf{0.353}&\textbf{0.236}&\textbf{0.148}&\textbf{0.662}\\
\bottomrule
\end{tabular}}
\label{tab:ray_surface_dis_esti_gaussian}  \vspace{-0.3cm}
\end{table*}

\section{Experiments}
\label{sec:exp}

\begin{table*}[tb] \tabcolsep=0.15cm  
    \centering
    \caption{Quantitative results of 3D meshes reconstructed from estimated ray surface distances. Our model is trained on Scannet/++ dataset.}\vspace{-0.2cm}
      \resizebox{1\textwidth}{!}{
    \begin{tabular}{r|r|ccccccc|ccccccc}
        \toprule
         \multicolumn{2}{c|}{\makecell{}}  &\multicolumn{7}{c|}{\textit{test on} $\rightarrow$ ARKitScenes}&\multicolumn{7}{c}{\textit{test on} $\rightarrow$ ScanNet/ScanNet++}\\
       \cmidrule{3-16}
       \multicolumn{2}{c|}{\makecell{}} &Acc.$\downarrow$&Comp.$\downarrow$&Pre.$\uparrow$&Rec.$\uparrow$&C-L1$\downarrow$&NC$\uparrow$&F1$\uparrow$&Acc.$\downarrow$&Comp.$\downarrow$&Pre.$\uparrow$&Rec.$\uparrow$&C-L1$\downarrow$&NC$\uparrow$&F1$\uparrow$\\
       \midrule
       3DGS~\cite{Kerbl2023}&TSDF
       &0.111&0.219&0.305&0.247&0.165&0.622&0.268&0.175&0.160&0.274&0.290&0.168&0.650&0.270\\
       GOF~\cite{Yu2024a}& MT
       &0.135&0.575&0.335&0.294&0.355&0.571&0.314&0.165&0.173&0.372&0.385&0.169&0.632&0.368\\
       PGSR~\cite{Chen2024a}&TSDF
       &0.153&0.188&0.437&0.344&0.171&\textbf{0.691}&0.378&\textbf{0.113}&0.088&\textbf{0.527}&0.543&\textbf{0.100}&\textbf{0.746}&0.532\\
       \midrule
       \textbf{\nickname{} (Ours)}  & TSDF
       &\textbf{0.051}&\textbf{0.083}&\textbf{0.671}&\textbf{0.609}&\textbf{0.067}&0.689&\textbf{0.633}&0.151&\textbf{0.066}&\textbf{0.527}&\textbf{0.633}&0.109&0.705&\textbf{0.566}\\
    \bottomrule
    \end{tabular}}
    \label{tab:mesh_eval} \vspace{-0.2cm}
\end{table*}

\begin{table*}[t]\tabcolsep=0.12cm 
\centering
\caption{Quantitative results of estimated ray-surface distances from 3D point clouds.}\vspace{-0.2cm}
  \resizebox{1\textwidth}{!}{
\begin{tabular}{@{}r|c|ccccc|ccccc|ccccc@{}}
    \toprule
   \multicolumn{2}{c|}{\makecell{}}&\multicolumn{5}{c|}{\textit{test on} $\rightarrow$ ARKitScenes}&\multicolumn{5}{c|}{\makecell{\textit{test on} $\rightarrow$ ScanNet/ScanNet++}}&\multicolumn{5}{c}{\makecell{\textit{test on} $\rightarrow$ MultiScan}}\\
   \cmidrule{3-17}
   \multicolumn{2}{c|}{\makecell{}}&ADE$\downarrow$ &RMSE$\downarrow$ &Abs-Rel$\downarrow$&Sq-Rel$\downarrow$&$\delta\uparrow $&ADE$\downarrow$&RMSE$\downarrow$&Abs-Rel$\downarrow$&Sq-Rel$\downarrow$&$\delta\uparrow$&
   ADE$\downarrow$&RMSE$\downarrow$&Abs-Rel$\downarrow$&Sq-Rel$\downarrow$&$\delta\uparrow$ \\
   \midrule
   & \multirow{4}{*}{\rotatebox{90}{\makecell{\small \textit{train on} \\ \small ARKitScenes}}} & \\
   Pointersect~\cite{Chang2023}&
   &0.335&0.456&0.198&0.132&0.643& 0.389&0.553&0.177&0.148&0.680& 0.254&0.330&0.201&0.093&0.637 \\
   RayDF~\cite{Liu2023a}&
   &0.166&0.281&0.112&0.074&0.876 &0.186&0.321&0.094&0.055&0.883    &0.154&0.234&0.134&0.062&0.822\\
   \textbf{\nickname{} (Ours)} &
   &\textbf{0.088}&\textbf{0.205}&\textbf{0.066}&\textbf{0.060}&\textbf{0.949}& \textbf{0.107}&\textbf{0.269}&\textbf{0.051}&\textbf{0.049}&\textbf{0.943} &\textbf{0.067}&\textbf{0.149}&\textbf{0.064}&\textbf{0.037}&\textbf{0.943}\\
   \midrule
   & \multirow{4}{*}{\rotatebox{90}{\makecell{\small \textit{train on} \\ \small ScanNet/++}}} & \\
    Pointersect~\cite{Chang2023}&
    &0.299&0.398&0.209&0.141&0.691 &0.344&0.477&0.181&0.133&0.708 &0.233&0.289&0.225&0.103&0.656\\
   RayDF~\cite{Liu2023a}&
   &0.289&0.407&0.189&0.118&0.699 &0.161&0.291&0.086&0.049&0.893 &0.349&0.429&0.313&0.259&0.561\\
   \textbf{\nickname{} (Ours)} & 
   &\textbf{0.096}&\textbf{0.217}&\textbf{0.074}&\textbf{0.077}&\textbf{0.946} &\textbf{0.093}&\textbf{0.234}&\textbf{0.047}&\textbf{0.040}&\textbf{0.949 }&\textbf{0.130}&\textbf{0.229}&\textbf{0.131}&\textbf{0.216}&\textbf{0.914}\\
\bottomrule
\end{tabular}}
\label{tab:exp_ray_surface_dis_esti_pc} \vspace{-0.3cm}
\end{table*}

\phantom{xxx}\textbf{Datasets}: Our method is evaluated on four real-world datasets based on the available train/test splits: \textbf{1) ScanNet} \cite{Dai2017} consisting of 1201 and 100 scenes for training and test; \textbf{2) ScanNet++} \cite{Yeshwanth2023} comprising 865 and 50 scenes for training and test; \textbf{3) ARKitScenes} \cite{Baruch2021} with 4498 and 549 scenes for training and test, which is the largest real-world indoor scene dataset captured by mobile RGBD sensor form Apple LiDAR scanner; \textbf{4) MultiScan} \cite{Mao2022} only consisting of about 200 indoor scenes in total captured by smartphones and tablets, where all scenes are used as a test set in our experiments. Due to the domain overlap, we merge ScanNet and ScanNet++, getting a joint train set and a joint test set.

\textbf{Baselines}: We choose 5 representative groups of methods as our baselines: 1) the state-of-the-art per-scene optimization based 3D Gaussians splatting methods \textbf{GOF} \cite{Yu2024a} and \textbf{PGSR} \cite{Chen2024a} particularly designed for high-fidelity surface reconstruction; 2) the latest feed-forward generalizable 3D Gaussian splatting method \textbf{MVSGaussian} \cite{Liu2024}; 
3) the state-of-the-art point cloud rendering method \textbf{PFGS} \cite{Wang2024a}; 4) the state-of-the-art ray-based methods \textbf{Pointersect} \cite{Chang2023} and \textbf{RayDF} \cite{Liu2023a} which is adapted as a generalizable version where the ray feature extractor is exactly the same as ours; 5) the latest image-based depth estimation methods \textbf{DepthAnythingV2} \cite{Yang2024} and \textbf{DepthPro} \cite{Bochkovskii2025}.

\textbf{Metrics}: Primarily, we evaluate the surface reconstruction by measuring the error between predicted ray-surface distance and ground truth distance. We report the per ray-surface \textbf{absolute distance error (ADE)} in meters across all test images, and other four commonly used metrics \cite{Turkulainen2025} including \textbf{Root Mean Squared Error (RMSE)}, the \textbf{Absolute Relative Distance (Abs Rel)}, the \textbf{Squared Relative Distance (Sq Rel)}, and the \textbf{Threshold accuracy ($\delta \leq 1.25$)} which measures the percentage of predicted value within a certain threshold $\delta$ of ground truth. In addition, we also evaluate the reconstructed 3D meshes, reporting \textbf{Accuracy}, \textbf{Completion}, \textbf{Precision}, \textbf{Recall}, \textbf{Chamfer-L1 distance}, \textbf{Normal Consistency}, and \textbf{F-scores} with a threshold of 5cm. 
All details are in Appendix~\ref{app_dataset}\&\ref{app_metric_ray}\&\ref{app_metric_mesh}\&\ref{app_baseline}.

\subsection{Evaluation on 3D Gaussians}\label{sec:exp_3d_gaussians}
In this setting, our method takes 3D Gaussians as input and predicts raylet distances for query rays. To prepare for this experiment, for every 3D scene in ScanNet/ScanNet++, ARKitScenes, and MultiScan, we first use the vanilla 3DGS to learn per-scene 3D Gaussisans from RGB images in an offline fashion, and then evaluate our method in the following two groups of experiments.
\begin{itemize}[leftmargin=*]
    \item Group 1: We train our method from scratch on the joint train set of ScanNet/ScanNet++. After that, we evaluate the trained model on the test sets of ScanNet/ScanNet++, ARKitScenes, and MultiScan.    
    \item Group 2: We train our method from scratch on the train set of ARKitScenes, and then test it on the test sets of ScanNet/ScanNet++, ARKitScenes, and MultiScan. 
\end{itemize} 

For a fair comparison, the four feed-forward generalizable baselines MVSGaussian \cite{Liu2024}, Pointersect \cite{Chang2023}, RayDF \cite{Liu2023a}, and PFGS \cite{Wang2024a} are all trained with the same amount of training data and supervision signals as our method. For a reference, the per-scene methods 3DGS \cite{Kerbl2023}, GOF \cite{Yu2024a} and PGSR \cite{Chen2024a} directly render depth images from their optimized 3D Gaussians on test set respectively. For the image-based depth estimation methods, we first render test view RGBs from 3D Gaussians and then feed them into a depth estimator, obtaining depth results. Lastly, we use ground truth depth values to align the scale of estimated depths, which is strongly in favor of these baselines. More details of experiments are in Appendix~\ref{app_baseline}. 

\begin{figure*}[!ht]
  \centering
  \includegraphics[width=1\linewidth]{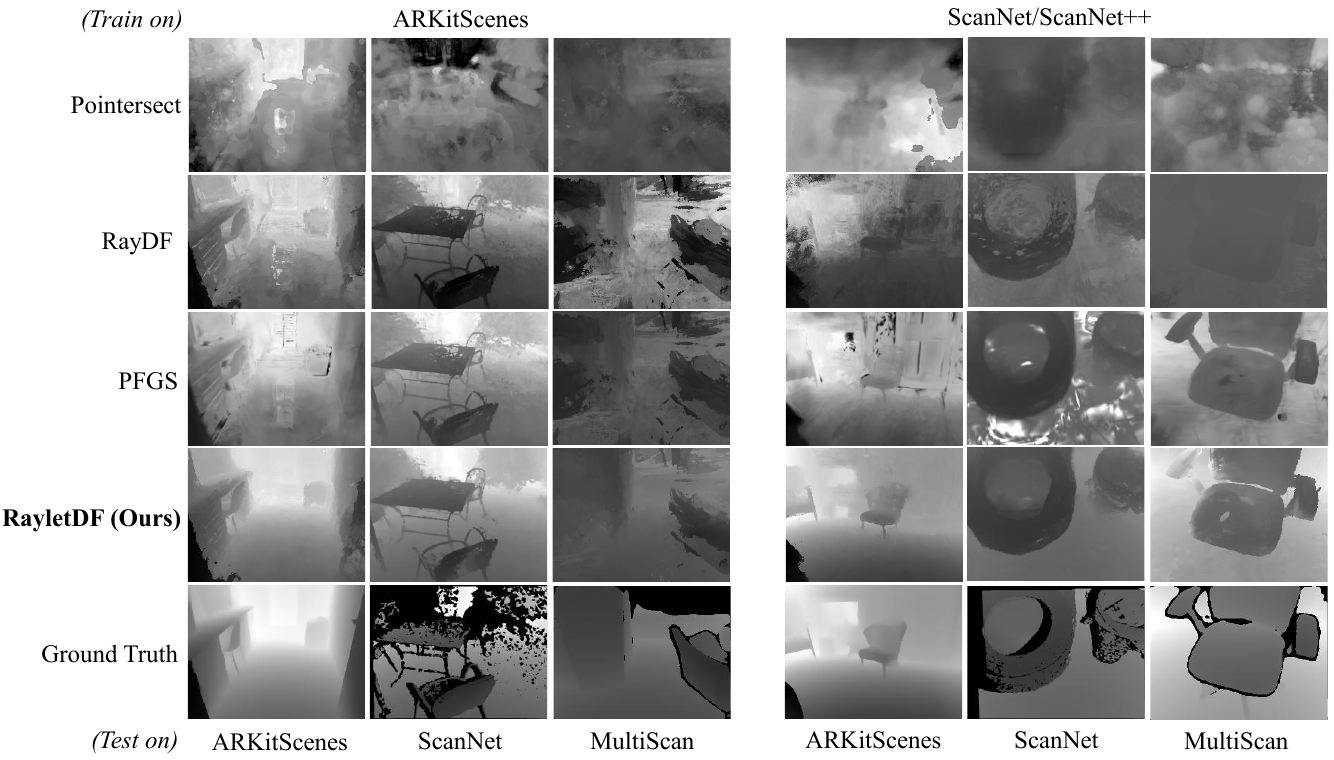}
  \centering
  \vspace{-0.2cm}
  \caption{Qualitative results of our method and baselines for 3D surface reconstruction on multiple datasets. All methods are trained on pre-estimated 3D Gaussians.}
  \label{fig:exp_qualitative_res_all_gs}
\end{figure*}

\textbf{Results \& Analysis}: Table \ref{tab:ray_surface_dis_esti_gaussian} compares the quantitative results of all methods for estimating distance values of all query rays from test views. Table \ref{tab:mesh_eval} compares the quantitative results of 3D meshes reconstructed by TSDF or Marching Tetrahedra (MT) from the rendered depth views for the three excellent per-scene Gaussian based methods and our method. Note that, in the original paper, GOF uses MT for higher quality surface reconstruction. Figure \ref{fig:exp_qualitative_res_all_gs} shows qualitative results. From the results, we can see that:
\begin{itemize}[leftmargin=*]
    \item When training/testing on ARKitScenes, ScanNet/ ScanNet++ datasets in domain, our method achieves the best accuracy, outperforming the second best method RayDF by large margins, with 0.115 $vs$ 0.183 meters on ARKitScenes and 0.145 $vs$ 0.202 meters on ScanNet/ScanNet++ on the key metric ADE. This means that our newly introduced raylet distance field has its clear advantage over existing ray-based representations.  
    \item When evaluating all methods across new datasets, our method demonstrates superior generalizability on unseen datasets, clearly surpassing all other baselines usually by more than 0.10 meters in accuracy. This performance gap is much greater than the in-domain reconstruction results. Essentially, this is because our learned raylet distance representations capture the local surface geometric patterns which tend to be generalizable at various scenes. 
    \item The per-scene optimization based 3D Gaussian methods and the depth estimation methods can achieve satisfactory results, but still being inferior than our method. 
\end{itemize}

\subsection{Evaluation on Point Clouds}\label{sec:exp_pointclouds}
In this setting, our method takes 3D point clouds as input and predicts raylet distances for query rays. To prepare for this experiment, for every 3D scene in ScanNet/ScanNet++, ARKitScenes, and MultiScan, we uniformly sample 10k points from the provided 3D scene mesh. Same as Section \ref{sec:exp_3d_gaussians}, we also conduct two groups of experiments. For a fair comparison, we also evaluate the top-performing ray-based methods Pointersect and RayDF with exactly the same training data and supervision signals as ours. 

\begin{figure*}[!ht]
  \centering
  \includegraphics[width=1\linewidth]{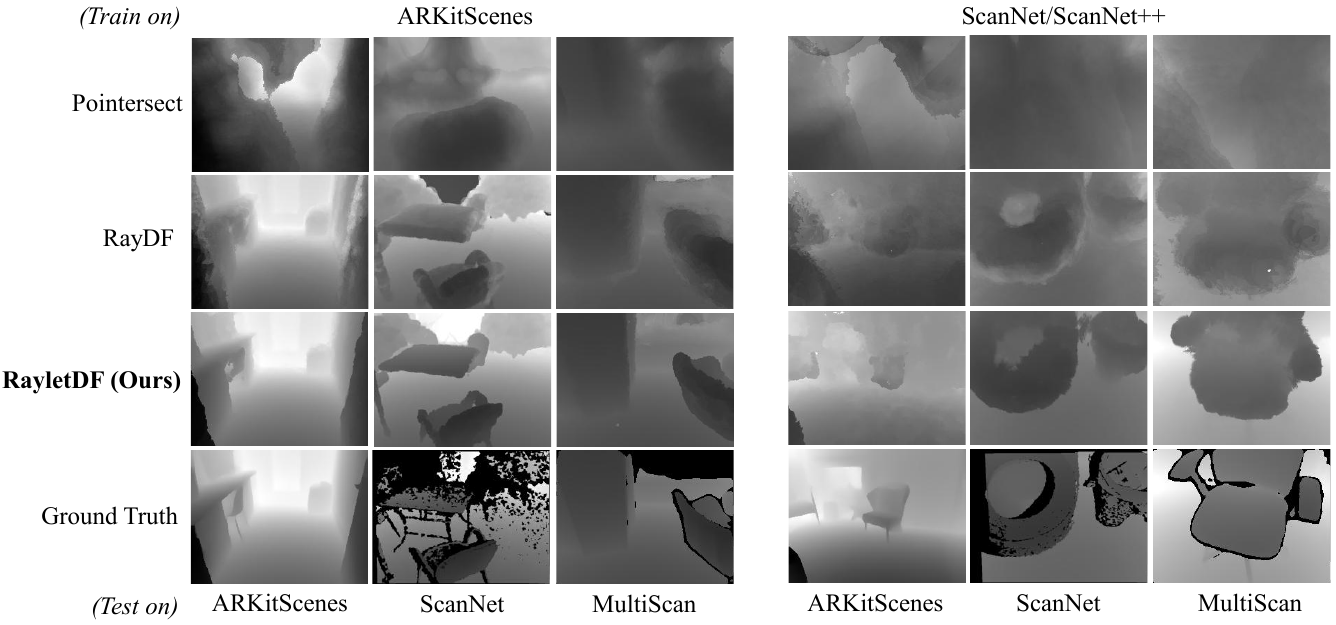}
  \centering
  \vspace{-0.2cm}
  \caption{Qualitative results of our method and baselines for 3D surface reconstruction on multiple datasets. All methods are trained on point clouds.}
  \label{fig:exp_qualitative_res_all_pc}
\end{figure*}

\textbf{Results \& Analysis}: Table \ref{tab:exp_ray_surface_dis_esti_pc} compares the quantitative results and Figure \ref{fig:exp_qualitative_res_all_pc} shows the qualitative results. We can see that, similar to the analysis in Section \ref{sec:exp_3d_gaussians}, our method demonstrates clearly better reconstruction accuracy on the in-domain datasets. Most notably, we achieve an outstanding generalization ability when evaluating on unseen datasets, clearly surpassing two strong baselines. For example, when trained on ARKitScenes or ScanNet/ScanNet++, our method achieves \{0.067, 0.130\} meters in ADE on the novel MultiScan dataset respectively, while the baselines often have more than 0.20 meters errors. Basically, our strong generalization ability on point clouds is also attributed to the effectiveness of our virtual ball based raylet sampling strategy designed in Section \ref{sec:sampling_raylets}. 

\subsection{Evaluation on Raylet Sampling in Testing}
Regarding our designed multi-raylet blender in Section \ref{sec:multiraylet_blender}, the single hyperparameter $T$ of this module can be different in training and test phase, allowing flexibility during evaluation on various datasets. Intuitively, once our model is well-trained on a dataset, in testing, the more number of raylet samples along every single ray, we may obtain more accurate surface estimations. 

We further conduct experiments to validate the flexibility of raylet sampling in test phase. In particular, we use our model well-trained on ScanNet/ScanNet++ where $T$ is chosen as 5 during training, and then directly evaluate on the test splits of ScanNet/ScanNet++, ARKitScenes, and the whole MultiScan dataset, where $T$ is chosen as $\{1, 5, 10, 20\}$ respectively. 

Tables \ref{tab:exp_raylet_sample_test_scan++}\&\ref{tab:exp_raylet_sample_test_arkit}\&\ref{tab:exp_raylet_sample_test_multiscan} show the quantitative results. It can be seen that, the more raylets sampled along every single ray during testing, we generally obtain more accurate surface reconstruction. This is particularly obvious when the trained model is evaluated on unseen datasets as shown in Tables \ref{tab:exp_raylet_sample_test_arkit}\&\ref{tab:exp_raylet_sample_test_multiscan}. Primarily, the is because if more raylets are sampled along every single light ray, our multi-raylet blender tends to vote a more accurate surface point, instead of being affected by outlier predictions among all estimations. This validates the generalizability and robustness of our simple design. More results are in Appendix \ref{app_sample_num_test}. 
\begin{table}[th] \vspace{-0.2cm}
\tabcolsep=0.15cm  
\centering
\caption{Qualitative results for different number of raylet samples per ray when evaluating on the test split of ScanNet/ScanNet++.}\vspace{-0.3cm}
\label{tab:exp_raylet_sample_test_scan++}
\resizebox{1.0\linewidth}{!}
{
\begin{tabular}{lccccc}
\toprule[1.0pt]
\textit{(train samples: $T=5$)}& ADE$\downarrow$ &RMSE$\downarrow$ &Abs-Rel$\downarrow$&Sq-Rel$\downarrow$&$\delta\uparrow $  \\
\toprule[1.0pt]

test samples: $T=1$ &0.177&0.337&0.087&0.190&0.896\\
test samples: $T=5$  &0.145&0.276&0.072&0.079&0.922\\
test samples: $T=10$ &\textbf{0.140}&\textbf{0.268}&\textbf{0.070}&\textbf{0.070}&\textbf{0.927}\\
test samples: $T=20$ &0.141&0.271&0.071&0.073&0.925\\

\toprule[1.0pt]
\end{tabular}
}
\vspace{-0.6cm}
\end{table}

\begin{table}[th] 
\tabcolsep=0.15cm  
\centering
\caption{Qualitative results for different number of raylet samples per ray when evaluating on the test split of ARKitScenes. }\vspace{-0.3cm}
\label{tab:exp_raylet_sample_test_arkit}
\resizebox{1.0\linewidth}{!}
{
\begin{tabular}{lccccc}
\toprule[1.0pt]
\textit{(train samples: $T=5$)} & ADE$\downarrow$ &RMSE$\downarrow$ &Abs-Rel$\downarrow$&Sq-Rel$\downarrow$&$\delta\uparrow $  \\
\toprule[1.0pt]

test samples: $T=1$ &0.161&0.287&0.113&0.076&0.876\\
test samples: $T=5$  &0.132&0.241&0.094&0.053&0.906\\
test samples: $T=10$ &0.127&0.234&0.092&\textbf{0.052}&0.912\\
test samples: $T=20$ &\textbf{0.124}&\textbf{0.230}&\textbf{0.090}&\textbf{0.052}&\textbf{0.917}\\

\toprule[1.0pt]
\end{tabular}
}
 \vspace{-0.6cm}
\end{table}

\begin{table}[th] 
\tabcolsep=0.15cm  
\centering
\caption{Qualitative results for different number of raylet samples per ray when evaluating on the whole MultiScan dataset.}\vspace{-0.3cm}
\label{tab:exp_raylet_sample_test_multiscan}
\resizebox{1.0\linewidth}{!}
{
\begin{tabular}{lccccc}
\toprule[1.0pt]
\textit{(train samples: $T=5$)} & ADE$\downarrow$ &RMSE$\downarrow$ &Abs-Rel$\downarrow$&Sq-Rel$\downarrow$&$\delta\uparrow $  \\
\toprule[1.0pt]

test samples: $T=1$ &0.291&0.393&0.261&0.177&0.619\\
test samples: $T=5$  &0.259&0.353&0.236&\textbf{0.148}&0.662\\
test samples: $T=10$ &0.252&\textbf{0.346}&\textbf{0.234}&\textbf{0.148}&0.674\\
test samples: $T=20$ &\textbf{0.247}&0.347&\textbf{0.234}&0.153&\textbf{0.687}\\

\toprule[1.0pt]
\end{tabular}
}
\vspace{-0.4cm}
\end{table}

\subsection{Ablations}
Our pipeline has three major modules, 1) Raylet Feature Extractor, 2) Raylet Distance Field, and 3) Multi-raylet Blender. To evaluate the effectiveness of each module and the sensitivity of hyperparameters, we conduct the following ablations on the merged ScanNet/ScanNet++ dataset, and the input to our method is choose as 3D Gaussians. 

\textbf{(1)$\sim$(3) Including different information of neighboring points}. Regarding the raylet feature extractor, in Equation \ref{eq:feat_neighbor_point}, for each neighboring point, we choose to include: (1) only the $xyz$ (\ie, $\boldsymbol{p}^k_l$) of a neighboring point, (2) only the relative position information \big(\ie, $(\frac{(\boldsymbol{p}_l^k - \boldsymbol{p}_l)}{||\boldsymbol{p}_l^k - \boldsymbol{p}_l ||} \oplus ||\boldsymbol{p}_l^k - \boldsymbol{p}_l ||$ \big), (3) both $xyz$ and the relative position information.

\textbf{(4)$\sim$(7) Choosing different number of neighboring points $K$}. For the raylet feature extractor, in Equation \ref{eq:feat_neighbor_point_No}, the hyperparameter $K$ is chosen as $\{1, 5, 10, 20\}$. This aims to evaluate the influence of local neighborhood on raylet features. In all main experiments, we choose $K$ as 5.

\textbf{(8)$\sim$(11) Sampling different number of raylets}. For the multi-raylet blender, along each light ray, a total of $T$ raylets are sampled near the surface. In this ablation, the hyperparameter $T$ is chosen as $\{1, 5, 10, 20\}$. In all main experiments, we choose $T$ as 5.

\textbf{(12) Removing the prediction of confidence score $s_l$}. Regarding the raylet distance field in Equation \ref{eq:raylet_dis_field}, we remove the confidence score, and the subsequent multi-raylet blender turns to simply average out multiple raylet distance predictions, \ie, $\hat{s}_{l_t}$ is $1/T$ in Equation \ref{eq:multi_raylet_blender}. 

\textbf{(13) Using alpha blending to aggregate multi-raylet predictions}. For the multi-raylet blender, we choose to regard the confidence score $s_l$ in Equation \ref{eq:raylet_dis_field} as opacity, and then use alpha blending to get the final prediction. 

\textbf{(14) Using sigmoid to normalize the confidence score $s_l$}. For the multi-raylet blender in Equation \ref{eq:multi_raylet_blender}, we turn to use sigmoid function to independently normalize confidence scores, meaning that multiple raylets along a single ray will be trained independently. 

Table \ref{tab:exp_abltaion_main} shows the ablation results. We can see that: 1) The greatest impact is caused by the decrease of raylet sample $T$ to be just 1 in training. Basically, this means that our raylet distance field degenerates to the existing ray-based method such as RayDF, which is ineffective to learn local surface patterns. 2) The second greatest impact is caused by the removal of confidence score $s_l$ in Equation \ref{eq:raylet_dis_field}. Basically, this is closely related to our design of multi-raylet blender as ablated in (12)$\sim$(14). Without a suitable multi-raylet blending strategy like ours, the accuracy of estimated surface drops. 3) Different number of neighboring points $K$ and choices of their features included could be helpful as shown in (1)$\sim$(7), but they are less important overall.   
\begin{table}[th] \vspace{-0.3cm}
\tabcolsep=0.12cm  
\centering
\caption{Results of all ablated models on ScanNet/ScanNet++. The main settings are bolded.}\vspace{-0.2cm}
\label{tab:exp_abltaion_main}
\resizebox{1.0\linewidth}{!}
{
\begin{tabular}{clccccc}
\toprule[1.0pt]
& & ADE$\downarrow$ &RMSE$\downarrow$ &Abs-Rel$\downarrow$&Sq-Rel$\downarrow$&$\delta\uparrow $  \\
\toprule[1.0pt]
(1) & Only $xyz$  &0.148&0.278&0.073&0.074&0.921\\
(2) & Only $\frac{(\boldsymbol{p}_l^k - \boldsymbol{p}_l)}{||\boldsymbol{p}_l^k - \boldsymbol{p}_l ||} \oplus ||\boldsymbol{p}_l^k - \boldsymbol{p}_l ||$ &0.146&0.275&0.072&0.078&0.921 \\
\textbf{(3)} & \textbf{Both} &\textbf{0.145}&\textbf{0.276}&\textbf{0.072}&\textbf{0.079}&\textbf{0.922}\\
\toprule[1.0pt]
(4) & $K=1$ &0.151&0.279&0.077&0.077&0.919\\
\textbf{(5)} & $\boldsymbol{K=5}$  &\textbf{0.145}&\textbf{0.276}&\textbf{0.072}&\textbf{0.079}&\textbf{0.922}\\
(6) & $K=10$ &0.150&0.280&0.074&0.074&0.920\\
(7) & $K=20$ &0.152&0.285&0.075&0.073&0.915\\
\toprule[1.0pt]
(8) & $T=1$ &0.174&0.326&0.085&0.144&0.897\\
\textbf{(9)} & $\boldsymbol{T=5}$  &\textbf{0.145}&\textbf{0.276}&\textbf{0.072}&\textbf{0.079}&\textbf{0.922}\\
(10) & $T=10$ &0.141&0.274&0.070&0.066&0.925\\
(11)&  $T=20$ &0.141&0.271&0.070&0.070&0.924\\
\toprule[1.0pt]
(12) & Removing score $s_l$ &0.170&0.302&0.084&0.097&0.907\\
(13) & Using alpha blending &0.156&0.290&0.075&0.073&0.914\\
(14)&  Using sigmoid &0.164&0.287&0.075&0.078&0.915\\
\textbf{(15)} & \textbf{The Full Pipeline (\nickname{})} &\textbf{0.145}&\textbf{0.276}&\textbf{0.072}&\textbf{0.079}&\textbf{0.922}\\
\toprule[1.0pt]
\end{tabular}
}
\vspace{-0.4cm}
\end{table}

\section{Conclusion}
\label{sec:conclusion}
In this paper, we present a generalizable 3D surface reconstruction method. By introducing a new technique of raylet distance field, our pipeline accurately captures intricate local surface patterns from both point clouds and 3D Gaussians, demonstrating superior performance across diverse real-world 3D scene datasets. Remarkably, thanks to the learned local raylet features, it exhibits excellent generalizability to new and unseen scenes in testing, while all baselines fail to do so.

\clearpage

\clearpage
\noindent\textbf{Acknowledgment:} 
This work was supported in part by Research Grants Council of Hong Kong under Grants 15225522 \& 25207822 \& 15219125, in part by Otto Poon Charitable Foundation Smart Cities Research Institute (8-CDCQ), in part by Research Centre for Unmanned Autonomous Systems (1-CE3D), The Hong Kong Polytechnic University. 

{
\small
\bibliographystyle{ieeenat_fullname}
\bibliography{references}
}

\clearpage
\setcounter{page}{1}
\maketitlesupplementary

\noindent\textbf{The appendix includes:}

\begin{itemize}
    \item Details for all the datasets.
    \item The definitions of all the evaluation metrics. 
    \item Implementation details for all the baselines.
    \item Implementation details for our model, including Raylet Feature Extractor, Raylet Distance Field Network, and the calculation of ray-gaussian intersection.
    \item Details for the surface normal derivation.
    \item Computation cost analysis
    \item Results on different point density
    \item More quantitative and qualitative results. \\
\end{itemize}

\subsection{Datasets}
\label{app_dataset}
\subsubsection{ScanNet++}
We use the official training and test splits from ScanNet++ dataset for novel view synthesis. There are 855 scenes for training and 50 scenes for test. For each scene in train set, we uniformly sample every 5th frame from the video sequence for training, and reserve every 10th frame in the test set for testing. We use the iPhone sequences with COLMAP registered poses and initial sparse points for pre-estimating 3D Gaussians. We do not use additional control over the number of Gaussian points during the training process.
\subsubsection{ScanNet}
We use the official training and test splits of ScanNetV2 dataset, with 1201 training scenes and 100 test scenes. Similar to ScanNet++, we also load every 5th frame from the trajectories for training, and then reserve every 10th frame in test set for evaluation. 
\subsubsection{ARKitScenes}
We use the official split of ``3dod" parts from ARKitScenes including 5047 rooms in total. After removing several rooms which cannot be pre-estimated by 3DGS, the training set contains 4496 rooms while the test dataset has 549 rooms. For training views, we sample every 5th image from the train set for training, and then reserve every 10th image in the test set for testing.
\subsubsection{MultiScan}
We use the whole MultiScan dataset with 207 scenes for testing, after removing a few scenes which cannot be pre-estimated by 3DGS. We sample every 10th image for test.

\subsection{Ray-surface Distance Metrics}
\label{app_metric_ray}
We report a series of ray distance metrics including Absolute Distance Error (ADE), RMSE, Absolute Relative Distance (Abs-Rel), Squared Relative Distance (Sq Rel) and Threshold accuracy ($\delta < t$) to measure the error between the estimated $i^{th}$ ray-surface distance $t_i^{pred}$ and the ground truth distance $t_i^{gt}$. These metrics are calculated over $N$ samples and defined as following:

\noindent\textbf{Absolute Distance Error (ADE)}:
$$\frac{1}{N}\sum_{i=1}^N|t_i^{pred} - t_i^{gt}|$$
\noindent\textbf{RMSE}:
$$\sqrt{\frac{1}{N}\sum_{i=1}^N(t_i^{pred} - t_i^{gt})^2}$$
\noindent\textbf{Absolute Relative Distance (Abs Rel)}:
$$\frac{1}{N}\sum_{i=1}^N\frac{|t_i^{pred} - t_i^{gt}|}{t_i^{gt}}$$
\noindent\textbf{Squared Relative Distance (Sq Rel)}:
$$\frac{1}{N}\sum_{i=1}^N\frac{(t_i^{pred} - t_i^{gt})^2}{t_i^{gt}} $$
\noindent\textbf{Threshold accuracy, $\delta$}:
$$\frac{1}{N}\sum_{i=1}^N[\text{max}(\frac{t_i^{pred}}{t_i^{gt}}, \frac{t_i^{gt}}{t_i^{pred}})<\delta]$$

\subsection{Evaluation Metrics on Meshes}
\label{app_metric_mesh}
The evaluation metrics for comparing predicted meshes $\hat{P}$ and ground truth meshes $P$ are defined as follows:

\noindent\textbf{Accuracy}:
$$\frac{1}{|\hat{P}|}\sum_{\mathbf{\hat{p}} \in \hat{P}}(\text{min}_{\textbf{p}\in P}\|\textbf{p}-\mathbf{\hat{p}}\|)$$

\noindent\textbf{Completion}
$$\frac{1}{|P|}\sum_{\textbf{p}\in P}(\text{min}_{\mathbf{\hat{p}} \in \hat{P}}\|\textbf{p}-\mathbf{\hat{p}}\|)$$

\noindent\textbf{Chamber-L1}
$$\frac{\text{Accuracy}+\text{Completion}}{2}$$

\noindent\textbf{Normal Accuracy}
$$\frac{1}{|\hat{P}|}\sum_{\mathbf{\hat{p}} \in \hat{P}}(\mathbf{n_{\mathbf{p}}^T\mathbf{n}_\mathbf{\hat{p}}})\ s.t. \ \textbf{p}=\mathop{argmin}\limits_{\textbf{p}\in P} \|\mathbf{p}-\mathbf{\hat{p}}\|$$

\noindent\textbf{Normal Completion}
$$\frac{1}{|{P}|}\sum_{\mathbf{{p}} \in {P}}(\mathbf{n_{\mathbf{p}}^T\mathbf{n}_\mathbf{\hat{p}}})\ s.t. \ \hat{\textbf{p}}=\mathop{argmin}\limits_{\hat{\textbf{p}}\in \hat{P}} \|\mathbf{p}-\mathbf{\hat{p}}\|$$

\noindent\textbf{Normal Consistency}
$$\frac{\text{Normal Accuracy + Normal Completion}}{2}$$

\noindent\textbf{Precision}
$$\frac{1}{|P|}\sum_{\mathbf{\hat{p}} \in \hat{P}}(\text{min}_{\textbf{p}\in P}\|\textbf{p}-\mathbf{\hat{p}}\|) < 5cm$$

\noindent\textbf{Recall}
$$\frac{1}{|P|}\sum_{\textbf{p}\in P}(\text{min}_{\mathbf{\hat{p}} \in \hat{P}}\|\textbf{p}-\mathbf{\hat{p}}\|)<5cm$$

\noindent\textbf{F-score}
$$\frac{2\times\text{Precision}\times\text{Recall}}{\text{Precision + Recall}}$$

\subsection{Baselines}
\label{app_baseline}
\noindent\textbf{3DGS}: We use the official implementation of Gaussian splatting~\cite{Kerbl2023} and increase densification gradient threshold to 0.0005 on datasets used in this paper to remove floaters which reduce the render performance significantly. Each scene Gaussian model of ARKitScenes dataset is trained for 7K iterations to obtain better render performance while scenes of other datasets are trained for 30K iterations. Additionally, we modify the original CUDA kernel to support depth rendering and convert depth predictions to ray-surface distances according to camera intrinsics.

\noindent\textbf{GOF}: We use the official GOF~\cite{Yu2024a} source code. The densification gradient threshold and training iterations are the same with 3DGS models mentioned above.

\noindent\textbf{PGSR} We use the official PGSR~\cite{Chen2024a} source code for comparison. Similarly to the GOF and 3DGS training strategy, we also increase the densification gradient threshold to 0.0005. Gaussian models of ARKitScenes dataset are optimized for 20K iterations until its loss converges. 

\noindent\textbf{DepthanythingV2}:
We use the official depth estimation checkpoint with ViT-L backbone in the original paper to estimate depth $d^{pred}$ for RGB images rendered by 3DGS. Then we align the scale of estimation results with ground truth depth $d^{gt}$ by forcing their median and variance to be the same:
$$m^{pred} = \text{median}(d^{pred})$$
$$s^{pred} = \frac{1}{N}\sum_{i=1}^N|d^{pred}_i - m^{pred}|$$
$$d^{pred}_{norm} = \frac{d^{pred} - m^{pred}}{s^{pred}} $$
$$m^{gt} = \text{median}(d^{gt})$$
$$s^{gt} = \frac{1}{N}\sum_{i=1}^N|d^{gt}_i - m^{gt}|$$
$$d^{pred}_{align} = d^{pred}_{norm}*s^{gt} + m^{gt}$$

\noindent\textbf{Depth-Pro}:
We also use the official checkpoint to estimate depth values for rendered RGB images. Since the outputs of Depth-Pro are depth values in meter, we directly align the median and variance of estimation to ground truth depths.

\noindent\textbf{MVSGaussians}: We use the official MVSGaussians source code. For a fair comparison, we also add ground truth depth signals to supervise rendering results in training. During inference stage, in favor of the baseline, we use MVSGaussian model to get better initialization and then optimize the Gaussian per scene for 7000 iterations.

\noindent\textbf{Pointersect}: We use the official source code to re-train the Pointersect model. For a fair comparison, the same RGB-D images are used as supervision signals as ours.

\noindent\textbf{RayDF}: We implement a ray-surface distance field conditioned on the same local geometry features as ours to compare generalization ability fairly in this paper.

\noindent\textbf{PFGS}: We use the official implementation code. We re-train the network using the same depth supervision signals as ours. 

Notably, we use ground truth depth values to align the scale of estimated depths from \textbf{DepthanythingV2} and \textbf{Depth-Pro}, which is strongly in favor of these baselines. The ground truth depth is not available in real practice. Thus, we compare the performances of these two models aligned to the gaussian-rendered depth values, as shown in Table \ref{tab:ray_surface_dis_esti_pc_appendix}.

\begin{table*}[th]\tabcolsep=0.12cm 
\centering
\caption{Quantitative results of depth estimation models aligned with different method.}\vspace{-0.2cm}
  \resizebox{1\textwidth}{!}{
\begin{tabular}{@{}r|ccccc|ccccc|ccccc@{}}
    \toprule
   &\multicolumn{5}{c|}{\textit{test on} $\rightarrow$ ARKitScenes}&\multicolumn{5}{c|}{\makecell{\textit{test on} $\rightarrow$ ScanNet/ScanNet++}}&\multicolumn{5}{c}{\makecell{\textit{test on} $\rightarrow$ MultiScan}}\\
   \midrule
   &ADE$\downarrow$ &RMSE$\downarrow$ &Abs-Rel$\downarrow$&Sq-Rel$\downarrow$&$\delta\uparrow $&ADE$\downarrow$&RMSE$\downarrow$&Abs-Rel$\downarrow$&Sq-Rel$\downarrow$&$\delta\uparrow$&
   ADE$\downarrow$&RMS$\downarrow$E&Abs-Rel$\downarrow$&Sq-Rel$\downarrow$&$\delta\uparrow$ \\
   \midrule
   DepthAnythingV2$^\dagger$~\cite{Yang2024}
   &0.206&0.294&0.144&0.093&0.860& 0.168&0.248& 0.118&0.066&0.887 &0.228&0.303&0.205&0.117&0.775\\
   Depth-Pro$^\dagger$~\cite{Bochkovskii2025}
   &0.294&0.403&0.202&0.242&0.750& 0.220&0.319&0.120&0.106&0.861 &0.280&0.363&0.234&0.157&0.659 \\
    \midrule
    DepthAnythingV2~\cite{Yang2024}
   &0.325&0.419&0.215&0.135&0.637& 1.022&1.267& 0.580&1.053&0.303 &0.460&0.538&0.347&0.237&0.352\\
   Depth-Pro~\cite{Bochkovskii2025}
   &0.379&0.495&0.246&0.225&0.594& 0.374&0.478&0.187&0.163&0.712 &0.483&0.571&0.361&0.258&0.333 \\

\bottomrule
\end{tabular}}
\label{tab:ray_surface_dis_esti_pc_appendix}
\end{table*}

\subsection{Details of Raylet Feature Extractor}
\label{app_spconv}
\begin{figure}[!ht]
  \centering
  \includegraphics[width=0.5\linewidth]{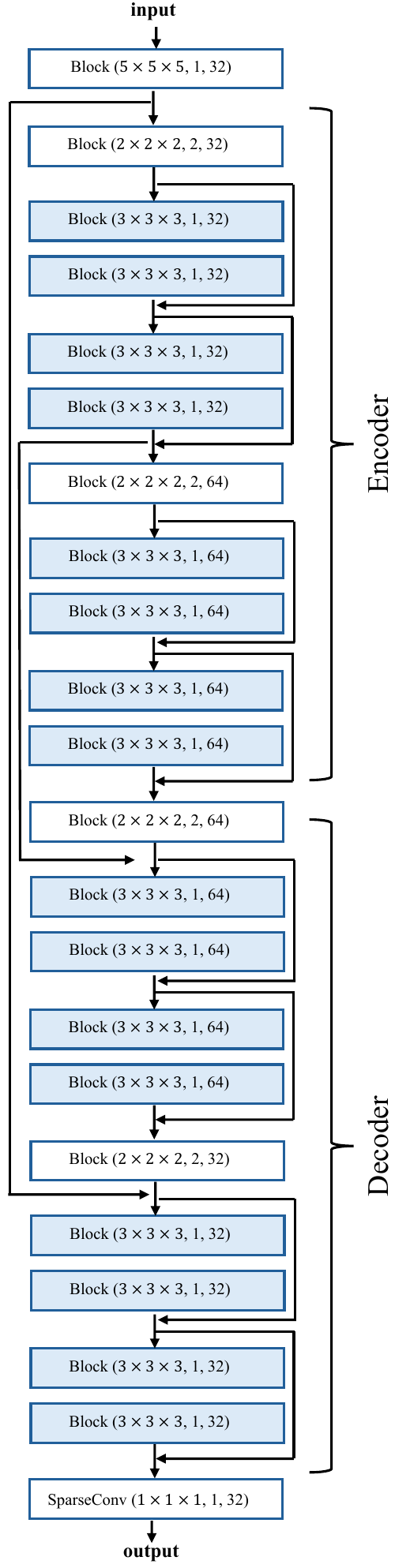}
  \centering
  \vspace{-0.2cm}
  \caption{Details of U-Net feature  extractor. Block means a sequential application of SparseConvolution in encoder or  SparseInverseConvolution in decoder, GroupNorm and ReLU activation.}
  \label{fig:spconv}
\end{figure}

We simply employ the SparseConv architecture as our Raylet Feature extractor. Particularly, we use the existing U-Net implementation of SPconv PyTorch package \url{https://github.com/traveller59/spconv}. As illustrated in Figure~\ref{fig:spconv}, the encoder and decoder have 8 layers respectively, which produce 32-dimensional features. Block means a sequential application of SparseConvolution in encoder or  SparseInverseConvolution in decoder, GroupNorm and ReLU activation.

\subsection{Details of Raylet Distance Field Network}
\label{app_raylet_mlp}
\begin{figure}[!ht]
  \centering
  \includegraphics[width=1.0\linewidth]{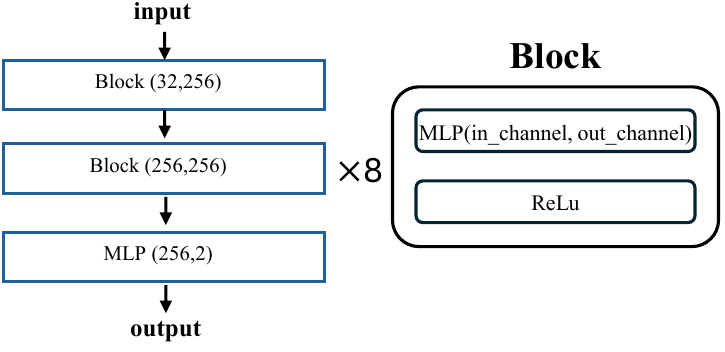}
  \centering
  \vspace{-0.2cm}
  \caption{Details Raylet Distance Field Network.}
  \label{fig:raylet_mlp}
\end{figure}

As shown in~\ref{fig:raylet_mlp}, we first use an MLP layer with ReLU to map the input feature embeddings to 256 dimensions, and then we use 8 dense layers of 256 neurons with ReLU activation network. Lastly, we use a linear layer with no activation function to output the raylet distance and confidence score.

\subsection{Calculation of Intersection}
\label{app_cal_intersect}
We provide the details of calculating intersections using Gaussians as an example.
Given a ray $x = r_o + t*r_d$ and a Gaussian $G(x)=exp(-\frac{1}{2}(x-\mu)^T\Sigma^{-1}(x-\mu))$, we can compute the Gaussian value on the ray as:
\begin{equation}
    G^{1D}(t) = exp(-\frac{1}{2}(o+t*r_d-\mu)^T\Sigma^{-1}(o+t*r_d-\mu))
\end{equation}
Follow~\cite{Keselman2022}, the intersection is defined as the point that maximizes the 1D Gaussian $G^{1D}(t)$. The distance $t_i$ can be calculated in closed-form by making the derivative of function $G^{1D}(t)$ equal to zero as:
\begin{equation}
    t_i = \frac{r_d\Sigma^{-1}(\mu-r_o)}{r_d^T\Sigma^{-1}r_d}
\end{equation}
Lastly, bringing $t_i$ back to ray function, we can obtain the intersection coordinates $x=r_o+t_i*r_d$.

To utilize the tile-based splitting strategy in 3DGS, we have customized the CUDA kernels to select the top K intersection points with the highest alpha blending weight.

\subsection{Derivation of Surface Normal}
\label{app_normal}
 In this section,we derive the formula of surface normal from our raylet distance field using a single raylet sample as an example.
 Given a ray from camera origin $o$, we can first parameter its direction in spherical coordinate system as $(\theta, \phi, 1)$, we convert the ray direction to world coordinate as $d = (x,y,z) = (\sin{\theta}\cos{\phi}, \cos{\theta}, \sin{\theta}\sin{\phi})$. Then we use the ray direction and origin to compute intersection between the Gaussians or points analytically and obtain the ray distance $t_{i}$ from camera origin to intersection point. We form a raylet $\bm{l}$ at the sample point position and use our network to predict the raylet distance. The complete ray distance from camera to surface is $D = F(\bm{l})+t_{i}$. Due to the differentiability of network and intersection calculations, we can calculate the derivative of surface point with respect to $\theta, \phi$ and the unit normal can be computed similarly as in~\cite{Liu2023a}. Specifically, the surface point coordinate can be formed as: 
 \begin{equation}
     \Phi(\theta, \phi) = (D\sin{\theta}\cos{\phi}, D\cos{\theta}, D\sin{\theta}\sin{\phi})
     \label{surface_pts}
 \end{equation}

From Eq. \eqref{surface_pts}, we have:
\begin{equation}
\frac{\partial \Phi}{\partial \phi} = 
\begin{bmatrix}
\left( \frac{\partial D}{\partial \phi} \cos \phi - D \sin \phi \right) \sin \theta \\
\frac{\partial D}{\partial \phi} \cos \theta\\
\left( \frac{\partial D}{\partial \phi} \sin \phi + D \cos \phi \right) \sin \theta 

\end{bmatrix}
\end{equation}

\begin{equation}
\frac{\partial \Phi}{\partial \theta} = 
\begin{bmatrix}
\left( \frac{\partial D}{\partial \theta} \sin \theta + D \cos \theta \right) \cos \phi \\
\frac{\partial D}{\partial \theta} \cos \theta - D \sin \theta\\
\left( \frac{\partial D}{\partial \theta} \sin \theta + D \cos \theta \right) \sin \phi \\

\end{bmatrix}
\end{equation}
Finally, the formula for a unit normal vector is :
\begin{equation}
    \mathbf{n} = \frac{\frac{\partial \Phi}{\partial \phi} \times \frac{\partial \Phi}{\partial \theta}}{\left\| \frac{\partial \Phi}{\partial \phi} \times \frac{\partial \Phi}{\partial \theta} \right\|}.
\end{equation}

\subsection{Quantitative and Qualitative Results on Normal Derivation}
\begin{figure}[h]
\centering
\caption{{Qualitative results of derived normals.}}
\vspace{-0.2cm}
\includegraphics[width=1.0\linewidth]{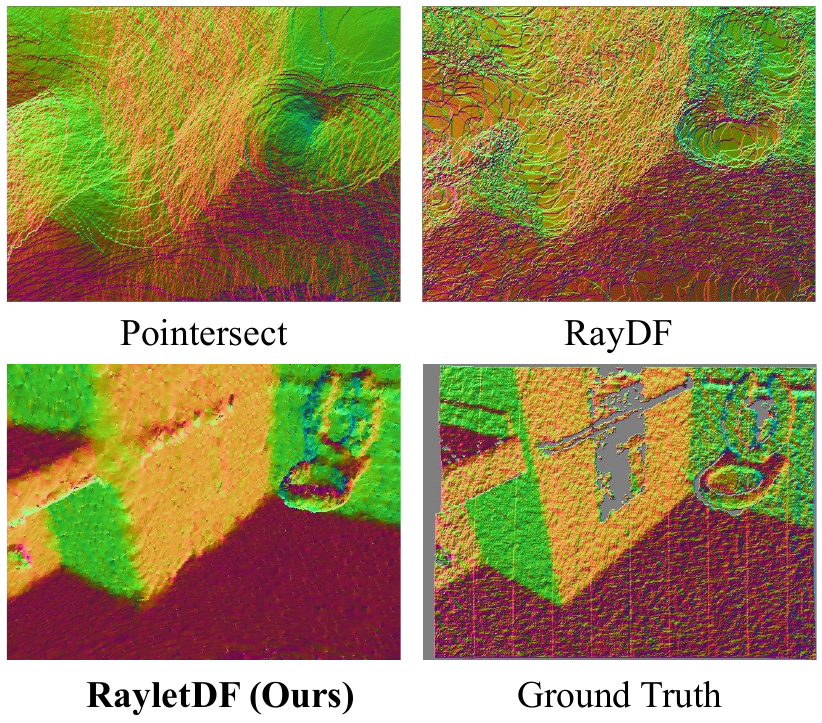}
\label{fig:normal_rebuttal}
\end{figure}

We report derived normals on sparse point cloud of ScanNet/++ test split in Table \ref{tab:normal} and Figure \ref{fig:normal_rebuttal}. Our method demonstrates superior performance compared to Pointersect and RayDF under this setting. Notably, our framework does not incorporate normal supervision during training, for a fair comparison normals for Pointersect are computed post-hoc from predicted depth maps. In contrast to RayletDF, RayDF lacks explicit local geometry-aware features in its architecture, preventing analytical normal estimation from the network. Consequently, we also derive RayDF's normals via depth map post-processing for fair comparison. The results suggest that RayletDF’s ray-based feature encoding and geometric reasoning enable more accurate normal estimation without direct supervision, outperforming depth-derived normals from competing methods.
\begin{table}[h]
\centering
 \caption{{Comparison of derived normals. The Mean Angle Error ($^\circ$), RMSE, mean of L1 error and median of L1 error are reported.}}\vspace{-0.2cm}
    \resizebox{1\linewidth}{!}{
  \begin{tabular}{r|cccc}
    \toprule[1.0pt]
      & MAE$\downarrow$ & RMSE$\downarrow$ & Mean L1$\downarrow$ &  Median L1$\downarrow$ \\
    \toprule[1.0pt]
    Pointersect~\cite{Chang2023}& 62.37 & 0.60     &  0.46 & 0.37\\
    RayDF~\cite{Liu2023a}     & 62.81  &0.61     &0.47   & 0.38\\
    \textbf{\nickname{} (Ours)}     &\textbf{48.86}   &  \textbf{0.50}   &  \textbf{0.37} & \textbf{0.28}\\

    \toprule[1.0pt]
  \end{tabular}}

  \label{tab:normal}
\end{table}

\subsection{More Results of Raylet Sampling in Testing}
\label{app_sample_num_test}
 We train our model on ScanNet/ScanNet++ given $T=\{1,10,20\}$ during training, and then test the three models on ScanNet/ScanNet++ with $T=\{1,5, 10,20\}$ samples in testing. Results are provided in  Tables~\ref{tab:exp_raylet_sample_test_scan++_trainT1}\&\ref{tab:exp_raylet_sample_test_scan++_trainT10}\&\ref{tab:exp_raylet_sample_test_scan++_trainT20}.

\begin{table}[h] 
\tabcolsep=0.15cm  
\centering
\caption{Qualitative results for different number of raylet samples per ray when evaluating on the test split of ScanNet/ScanNet++}\vspace{-0.2cm}
\label{tab:exp_raylet_sample_test_scan++_trainT1}
\resizebox{1.0\linewidth}{!}
{
\begin{tabular}{lccccc}
\toprule[1.0pt]
&\multicolumn{5}{c}{\textit{test on} $\rightarrow$ ScanNet/ScanNet++}\\
\midrule
\textit{(train samples: T=1)} & ADE$\downarrow$ &RMSE$\downarrow$ &Abs-Rel$\downarrow$&Sq-Rel$\downarrow$&$\delta\uparrow $  \\
\toprule[1.0pt]

 test samples: $T=1$ &0.174&0.326&0.085&0.144&0.898\\
 test samples: $T=5$  &0.168&0.292&0.082&0.086&0.908\\
 test samples: $T=10$ &0.174&0.306&0.087&0.085&0.904\\
 test samples: $T=20$ &0.172&0.301&0.0852&0.0851&0.905\\

\toprule[1.0pt]
\end{tabular}
}
\end{table}

\begin{table}[h] 
\tabcolsep=0.15cm  
\centering
\caption{Qualitative results for different number of raylet samples per ray when evaluating on the test split of ScanNet/ScanNet++} \vspace{-0.2cm}
\label{tab:exp_raylet_sample_test_scan++_trainT10}
\resizebox{1.0\linewidth}{!}
{
\begin{tabular}{lccccc}
\toprule[1.0pt]
&\multicolumn{5}{c}{\textit{test on} $\rightarrow$ ScanNet/ScanNet++}\\
\midrule
\textit{(train samples: T=10)} & ADE$\downarrow$ &RMSE$\downarrow$ &Abs-Rel$\downarrow$&Sq-Rel$\downarrow$&$\delta\uparrow $  \\
\toprule[1.0pt]

 test samples: $T=1$ &0.190&0.357&0.093&0.184&0.884\\
 test samples: $T=5$  &0.146&0.278&0.072&0.076&0.920\\
 test samples: $T=10$ &0.141&0.274&0.070&0.066&0.925\\
 test samples: $T=20$ &0.142&0.275&0.071&0.069&0.924\\

\toprule[1.0pt]
\end{tabular}
}
\end{table}

\begin{table}[h] 
\tabcolsep=0.15cm  
\centering
\caption{Qualitative results for different number of raylet samples per ray when evaluating on the test split of ScanNet/ScanNet++}\vspace{-0.2cm}
\label{tab:exp_raylet_sample_test_scan++_trainT20}
\resizebox{1.0\linewidth}{!}
{
\begin{tabular}{lccccc}
\toprule[1.0pt]
&\multicolumn{5}{c}{\textit{test on} $\rightarrow$ ScanNet/ScanNet++}\\
\midrule
\textit{(train samples: T=20)} & ADE$\downarrow$ &RMSE$\downarrow$ &Abs-Rel$\downarrow$&Sq-Rel$\downarrow$&$\delta\uparrow $  \\
\toprule[1.0pt]

 test samples: $T=1$ &0.194&0.368&0.096&0.329&0.881\\
 test samples: $T=5$  &0.146&0.277&0.072&0.076&0.920\\
 test samples: $T=10$ &0.139&0.267&0.069&0.068&0.926\\
 test samples: $T=20$ &0.141&0.271&0.070&0.070&0.924\\

\toprule[1.0pt]
\end{tabular}
}
\end{table}

\begin{table*}[h] 
\centering
 \caption{{The time, memory consumption, and the speed of rendering a $640\times480$ image given different number of points sampled.}}\label{tab:speed} \vspace{-0.2cm}
\resizebox{1\linewidth}{!}{
\begin{tabular}{r|ccccc}
\toprule[1.0pt]
& Total Train Time (hours) & Memory (GB) & No. of Samples per ray & Render Speed (FPS) & ADE $\downarrow$\\
\toprule[1.0pt]
NeRF~\cite{Mildenhall2020}                     & $\sim$ 645 & 4.39 / 0.64 & 192 (dense) / 25 & 0.11 / 1.16  & - \\ 
InstantNGP~\cite{muller2022instant}               & $\sim$ 27.5 & 14.39 / 12.94 & 1024 (dense) / 25 & 1.09 / 1.17 & - \\ 
    NeuS~\cite{wang2021neus}                 & $\sim$ 469 & 6.20 / 1.22 & 160 (dense) / 25 & 0.06 / 0.17 & - \\ \hline
    3DGS~\cite{Kerbl2023}& $\sim$ 37.5 & 4.45&     - &293 & 0.321 \\ \hline 
    \nickname{} (Ours)   & $\sim$ (37.5+30)&  11.63&     5$\times$5 &5.35 & 0.145 \\ \hline
    \toprule[1.0pt]
  \end{tabular}
}
\end{table*}

\subsection{Computation Cost Analysis}

We report the total train time, memory consumption and rendering speed in test for coordinate-based methods, 3DGS and ours on ScanNet/++ test split in Table \ref{tab:speed}. Voxel grid KNN \url{https://github.com/janericlenssen/torch_knnquery/} is employed to accelerate our algorithm. Since our method only needs 5 raylets sampled on each ray and each raylet only has 5 neighboring points sampled, for a fair comparison, we also only sample 25 points for coordinate-based methods in addition to their original dense samples. For clarification, our training time includes 30 hours to train RayletDF on ScanNet/++ train split, and 37.5 hours of pre-estimating 3D Gaussians on ScanNet/++ test split. 
We can see that: 1) our method has a clear advantage over coordinate-based methods in rendering speed; 2) though being slower than 3DGS, our method achieves much higher accuracy in surface reconstruction.  

\subsection{Results of Pre-trained Models}
Results of pre-trained Pointersect on sparse point clouds are given in Table \ref{tab:pre-train-pt}. The model is pre-trained on dense point clouds, thus being inferior to our trained version. RayDF is a per-scene method and we cannot use its pre-trained models. 
\begin{table}[h]
  \centering
   \caption{{Results of pre-trained pointersect model.}}\vspace{-0.2cm}
  \setlength{\tabcolsep}{2pt}
      \resizebox{1\linewidth}{!}{
  \begin{tabular}{r|ccccc}
    \toprule
      &ADE$\downarrow$ &RMSE$\downarrow$ &Abs-Rel$\downarrow$&Sq-Rel$\downarrow$&$\delta\uparrow$ \\
    \midrule
    ARKitScenes       & 0.57 &1.14 &0.39&1.33&0.65\\
    ScanNet/ScanNet++ & 0.48 &0.97 & 0.23&0.69&0.77\\
    Multiscan         & 0.42 &0.89 & 0.31&0.96&0.71\\
    \bottomrule
  \end{tabular}
  }
  \label{tab:pre-train-pt}
\end{table}

\subsection{Quantitative Results on Different Point Density}
Given that Gaussian point clouds typically contain hundreds of thousands to millions of points per scene, we adopt sparse point cloud inputs in our primary experiments to rigorously evaluate the performance of RayletDF under low-density conditions. Furthermore, advancements in RGB image-based reconstruction techniques—such as Structure-from-Motion (SfM)—have made it increasingly practical to generate sparse point cloud data, aligning with our experimental design.

In this section, we test models on one million of points on ScanNet/++ test split in Table \ref{tab:1m_pts} to investigate the sensitivity to point density of our algorithm. Shown in the Table, our method is robust to high density points though only trained on sparse point settings. The pre-trained Pointersect model is originally trained on millions of points, we give its results for reference. As the input points already contain accurate surfaces, the performance gap between RayletDF and other methods decreases to some degree.

 \begin{table}[h]
  \centering
 \caption{{Results on one million points as input point clouds.}}\vspace{-0.2cm}
  \setlength{\tabcolsep}{2pt}
    \resizebox{1\linewidth}{!}{
  \begin{tabular}{l|ccccc}
    \toprule[1.0pt]
     &ADE$\downarrow$ &RMSE$\downarrow$ &Abs-Rel$\downarrow$&Sq-Rel$\downarrow$&$\delta\uparrow$ \\
   \toprule[1.0pt]
    Pre-trained Pointersect~\cite{Chang2023}   & 0.098 &0.264 &\textbf{0.040}&0.051& \textbf{0.958}  \\
    Pointersect~\cite{Chang2023}   & 0.197 & 0.354 &0.126&0.105& 0.844  \\
     RayDF~\cite{Liu2023a}    & 0.160 &0.271 &0.100&0.056&0.918\\
    \textbf{\nickname{} (Ours) }    & \textbf{0.088} &\textbf{0.232} &0.043&\textbf{0.043}&0.954\\

    \toprule[1.0pt]
  \end{tabular}}
  \label{tab:1m_pts}
\end{table}

\subsection{Ablation of Transformer Layer for Neighbor Points Aggregation}
 We present results of a Transformer layer to aggregate neighbor features on Gaussians ScanNet/++ test split in Table \ref{tab:pre-atten}. Employing transformer layer does not yield performance improvements in this setting, we hypothesize that much larger dataset is needed to fully unleash the ability of Transformer. 
\begin{table}[h]
  \centering
   \caption{{Comparison between Transformer layer and MLPs.}}\vspace{-0.2cm}
    \resizebox{1\linewidth}{!}{
  \begin{tabular}{r|ccccc}
    \toprule[1.0pt]
      &ADE$\downarrow$ &RMSE$\downarrow$ &Abs-Rel$\downarrow$&Sq-Rel$\downarrow$&$\delta\uparrow$ \\
    \toprule[1.0pt]
    Transformer       & 0.155 &0.286 &0.077&0.082&0.916\\
    MLPs             & 0.145 & 0.276 & 0.072  & 0.079  & 0.922\\
    \toprule[1.0pt]
  \end{tabular}}

  \label{tab:pre-atten}
\end{table}

 \begin{table}[!ht]
  \centering
\caption{Scene size and average distance (meter) between nearest neighbors in sparse point clouds.}
  \setlength{\tabcolsep}{2pt}
  \resizebox{1\linewidth}{!}{
  \begin{tabular}{l|ccccccc}
    \toprule
    Method &Mean X &Mean Y &Mean Z&Max X &Max Y &Max Z&Dist \\
    \midrule
    ARKitScenes       & 7.74 & 7.65 & 3.24 &22.23&17.98&9.15&0.08\\
    ScanNet/++        & 5.87 & 4.94 & 2.61 & 19.72  & 19.04&7.68&0.07\\
    Multiscan            & 6.32 & 2.89 & 6.36  & 16.49  & 5.36 &19.30 &0.07\\
    \bottomrule
  \end{tabular}}
 \vspace{-0.2cm}

  \label{tab:size}
\end{table}

\subsection{Standard Deviation Results}
We report standard deviation results on ray-based methods in Table \ref{tab:standard deviation} and Table \ref{tab:standard deviation pts}. Shown in the Table, our method achieves the most robust performance.

\begin{table*}[]\tabcolsep=0.1cm 
\centering
\caption{Standard deviation results on Gaussians dataset.}\vspace{-0.2cm}
\begin{tabular}{@{}r|c|cc|cc|cc@{}}
    \toprule
     \multicolumn{2}{c|}{\makecell{}} &\multicolumn{2}{c|}{\textit{test on} $\rightarrow$ ARKitScenes}&\multicolumn{2}{c|}{\makecell{\textit{test on} $\rightarrow$ ScanNet/ScanNet++}}&\multicolumn{2}{c}{\makecell{\textit{test on} $\rightarrow$ MultiScan}}\\
   \cmidrule{3-8}
   \multicolumn{2}{c|}{\makecell{}}&ADE $\downarrow$ &RMSE$\downarrow$ &ADE$\downarrow$&RMSE$\downarrow$&
   ADE$\downarrow$&RMSE$\downarrow$ \\
   \midrule
   Pointersect~\cite{Chang2023}&\multirow{3}{*}{\rotatebox{90}{\tiny \makecell{ \textit{train on} \\ ARKitScenes}}}
   &0.286$\pm$0.179&0.397$\pm$0.272      &0.366$\pm$0.184&0.259$\pm$0.381        &0.266$\pm$0.135  &0.311$\pm$0.176\\
   RayDF~\cite{Liu2023a}& &
   0.183$\pm$0.113&0.303$\pm$0.211       & 0.175$\pm$0.124&0.320$\pm$0.214       &0.326$\pm$0.183&0.425$\pm$0.233\\
   \textbf{\nickname{} (Ours)} & &
   \textbf{0.115$\pm$0.084} &\textbf{0.218$\pm$0.181}       &\textbf{0.175$\pm$0.123}&\textbf{0.320$\pm$0.217} 
    &\textbf{0.216$\pm$0.151}&\textbf{0.311$\pm$0.209}
   \\
   \midrule
    Pointersect~\cite{Chang2023}&\multirow{3}{*}{\rotatebox{90}{\tiny \makecell{\textit{train on} \\ ScanNet / ++}}}
           &0.328$\pm$0.253&0.462$\pm$0.351    &0.433$\pm$0.209&0.604$\pm$0.274              &0.404$\pm$0.215&0.504$\pm$0.249\\
   RayDF~\cite{Liu2023a}&  &0.587$\pm$0.276 &0.704$\pm$0.334   &0.202$\pm$0.120&0.337$\pm$0.198             &0.604$\pm$0.298 &0.690$\pm$0.327\\
   \textbf{\nickname{} (Ours)}  & 
    &\textbf{0.132$\pm$0.094}&\textbf{0.241$\pm$0.193}     &\textbf{0.145$\pm$0.100}&\textbf{0.276$\pm$0.186}     &\textbf{0.259$\pm$0.167} &\textbf{0.353$\pm$0.213}\\
\bottomrule
\end{tabular}
\label{tab:standard deviation} 
\end{table*}

\begin{table*}[]\tabcolsep=0.1cm 
\centering
\caption{Standard deviation results on point cloud dataset.}\vspace{-0.2cm}
\begin{tabular}{@{}r|c|cc|cc|cc@{}}
    \toprule
     \multicolumn{2}{c|}{\makecell{}} &\multicolumn{2}{c|}{\textit{test on} $\rightarrow$ ARKitScenes}&\multicolumn{2}{c|}{\makecell{\textit{test on} $\rightarrow$ ScanNet/ScanNet++}}&\multicolumn{2}{c}{\makecell{\textit{test on} $\rightarrow$ MultiScan}}\\
   \cmidrule{3-8}
   \multicolumn{2}{c|}{\makecell{}}&ADE $\downarrow$ &RMSE$\downarrow$ &ADE$\downarrow$&RMSE$\downarrow$&
   ADE$\downarrow$&RMSE$\downarrow$ \\
   \midrule
   Pointersect~\cite{Chang2023}&\multirow{3}{*}{\rotatebox{90}{\tiny \makecell{ \textit{train on} \\ ARKitScenes}}}
   &0.335$\pm$0.138&0.456$\pm$0.187      &0.389$\pm$0.193&0.553$\pm$0.278        &0.254$\pm$0.077  &0.330$\pm$0.099\\
   RayDF~\cite{Liu2023a}& &
   0.166$\pm$0.119&0.303$\pm$0.216       & 0.186$\pm$0.089&0.321$\pm$0.172       &0.154$\pm$0.098&0.234$\pm$0.152\\
   \textbf{\nickname{} (Ours)} & &
   \textbf{0.088$\pm$0.071} &\textbf{0.205$\pm$0.167}       &\textbf{0.107$\pm$0.068}&\textbf{0.269$\pm$0.164} 
    &\textbf{0.067$\pm$0.048}&\textbf{0.149$\pm$0.109}
   \\
   \midrule
    Pointersect~\cite{Chang2023}&\multirow{3}{*}{\rotatebox{90}{\tiny \makecell{\textit{train on} \\ ScanNet / ++}}}
           &0.299$\pm$0.115&0.398$\pm$0.156    &0.344$\pm$0.178&0.477$\pm$0.240             &0.233$\pm$0.086&0.289$\pm$0.096\\
   RayDF~\cite{Liu2023a}&  &0.289$\pm$0.178 &0.407$\pm$0.268   &0.161$\pm$0.085&0.291$\pm$0.158             &0.349$\pm$0.189 &0.429$\pm$0.222\\
   \textbf{\nickname{} (Ours)}  & 
    &\textbf{0.096$\pm$0.077}&\textbf{0.217$\pm$0.179}     &\textbf{0.093$\pm$0.056}&\textbf{0.234$\pm$0.133}     &\textbf{0.130$\pm$0.177} &\textbf{0.229$\pm$0.255}\\
\bottomrule
\end{tabular}
\label{tab:standard deviation pts} 
\end{table*}

\subsection{Scene Scale and Point Density Information}
We report the scale of scenes and distance to the nearest point of sparse point clouds used in our three group experiments in Table \ref{tab:size}.

\subsection{Additional Quantitative and Qualitative Results}
\begin{figure*}[ht]
  \centering
  \includegraphics[width=0.98\linewidth]{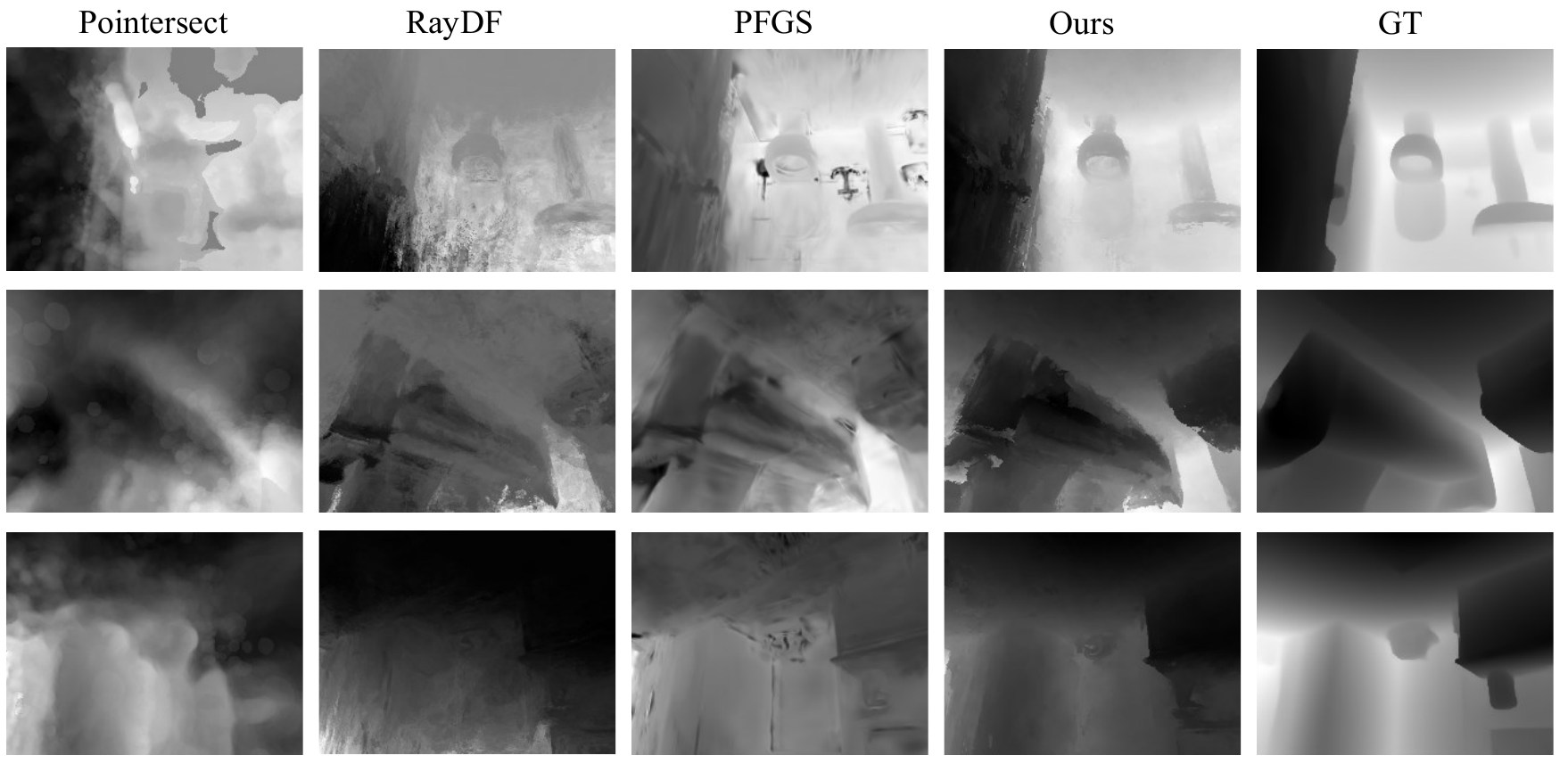}
  \centering
  \vspace{-0.2cm}
  \caption{Qualitative results of all methods trained on Gaussians of ScanNet/ScanNet++ and tested on ARKitScenes.}
  \label{fig:gs_scans2arkit}
\end{figure*}

\begin{figure*}[ht]
  \centering
  \includegraphics[width=0.98\linewidth]{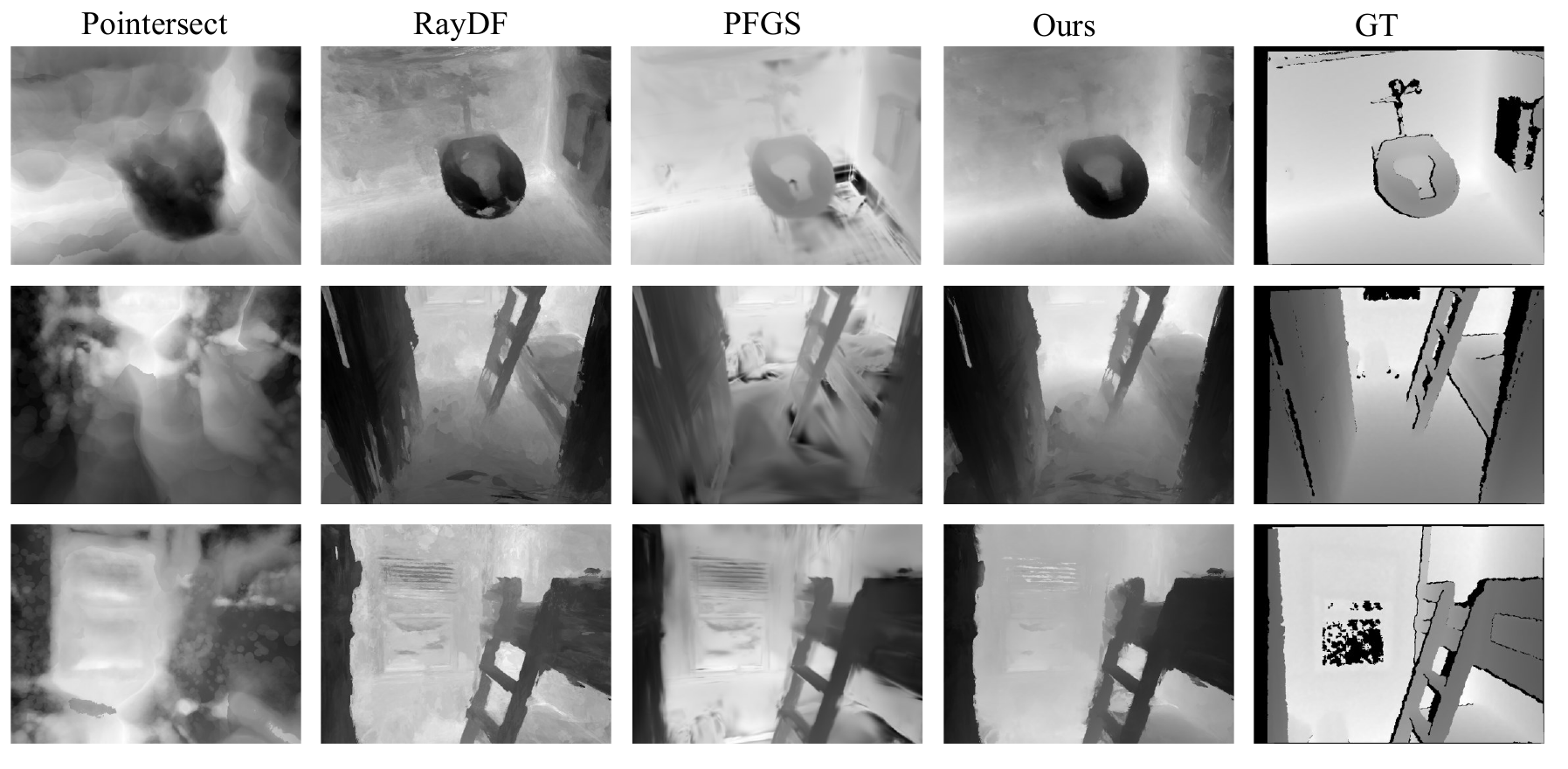}
  \centering
  \vspace{-0.2cm}
  \caption{Qualitative results of all methods trained on Gaussians of ScanNet/ScanNet++ and tested on ScanNet/ScanNet++.}
  \label{fig:gs_scans2scans}
\end{figure*}

More results are provided in the following table and figures. 
\textbf{Models trained on Gaussians of ScanNet/ScanNet++:} in Figure~\ref{fig:gs_scans2arkit}\&\ref{fig:gs_scans2scans}, we show more qualitative results of Group 1.

\begin{figure*}[hb]
  \centering
  \includegraphics[width=0.98\linewidth]{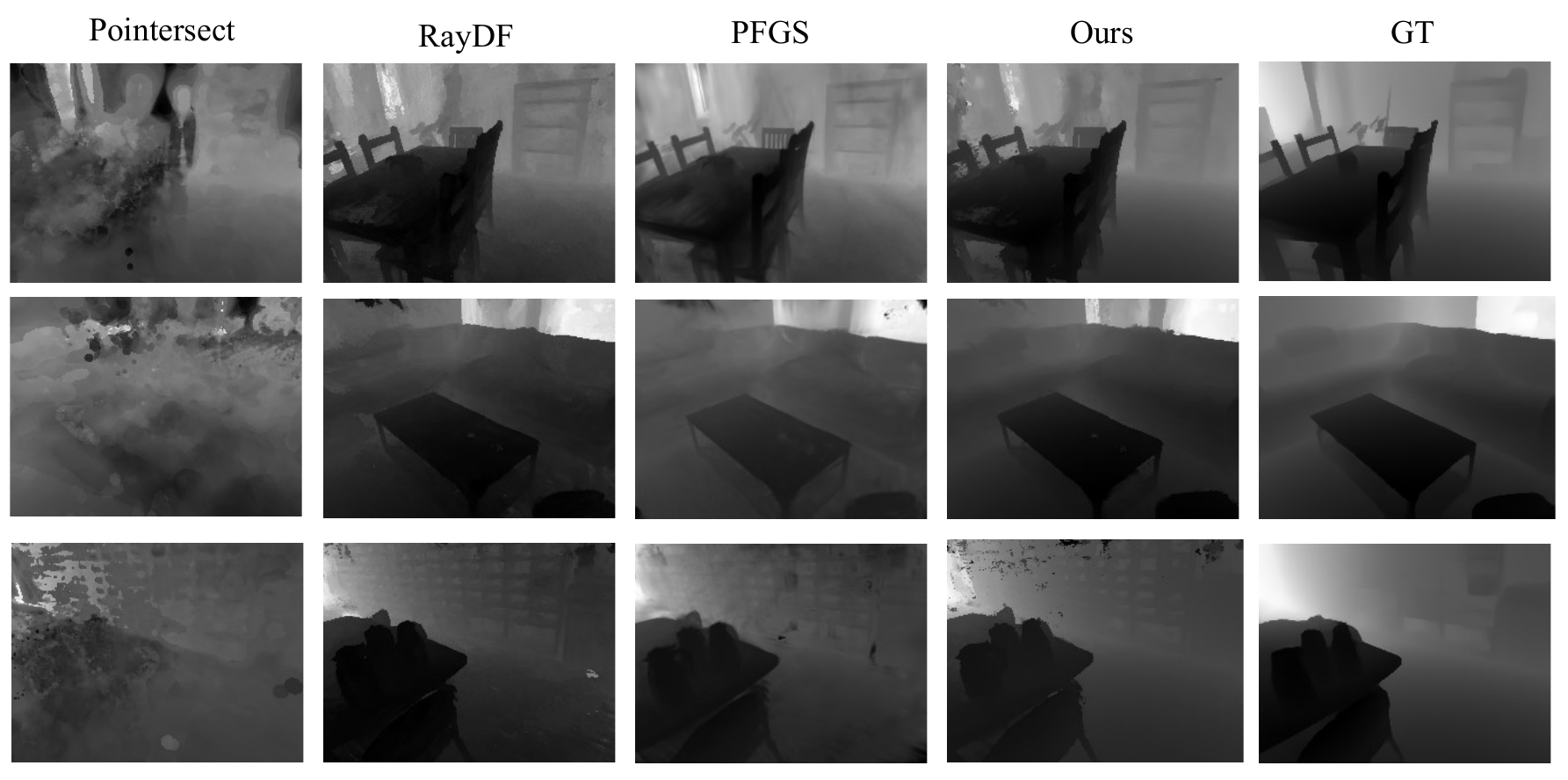}
  \centering
  \vspace{-0.2cm}
  \caption{Qualitative results of all methods trained on Gaussians of ARKitScenes and tested on ScanNet/ScanNet++.}
  \label{fig:gs_arkit2scans}
\end{figure*}

\begin{figure*}[ht]
  \centering
  \includegraphics[width=0.98\linewidth]{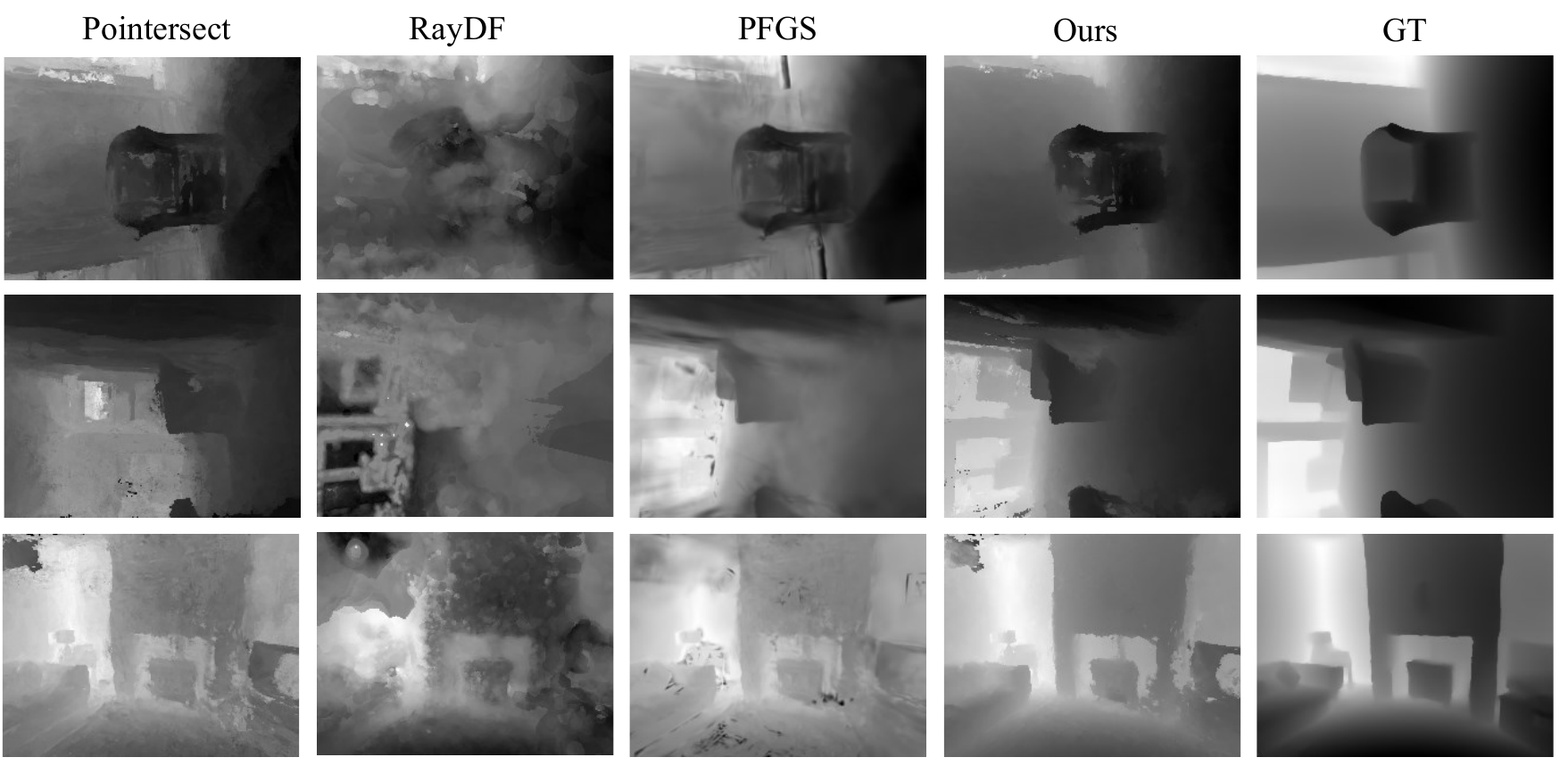}
  \centering
  \vspace{-0.2cm}
  \caption{Qualitative results of all methods trained on Gaussians of ARKitScenes and tested on ARKitScenes.}
  \label{fig:gs_arkit2arkit}
\end{figure*}

\begin{figure*}[ht]
  \centering
  \includegraphics[width=0.98\linewidth]{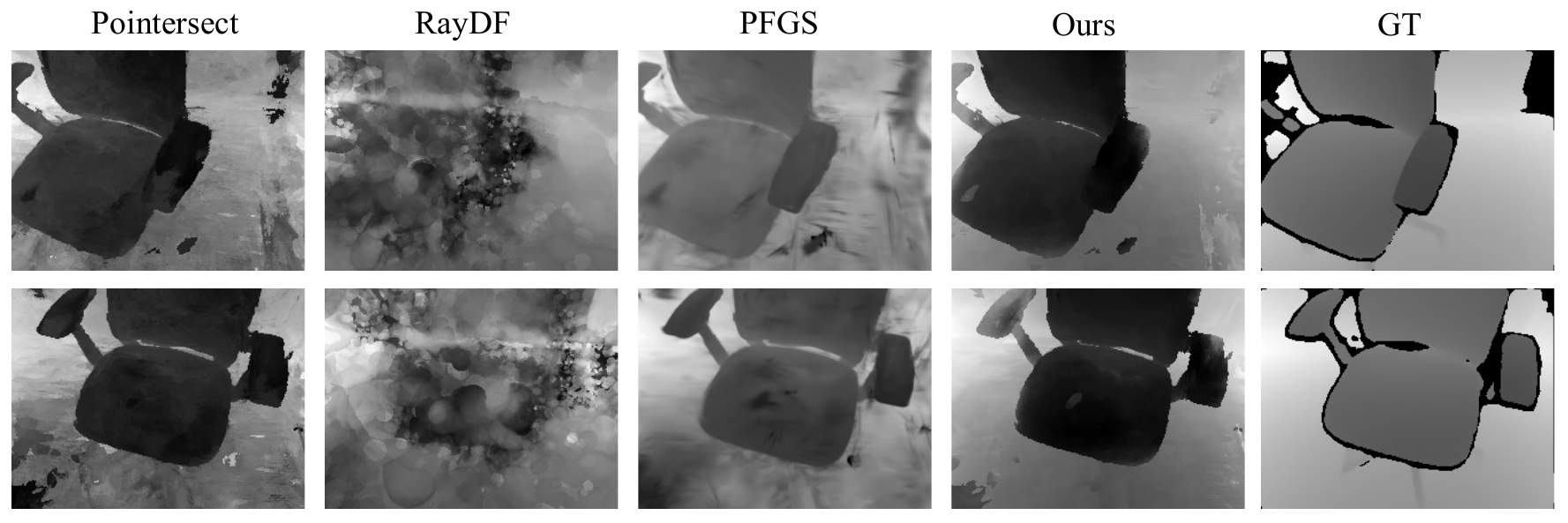}
  \centering
  \vspace{-0.2cm}
  \caption{Qualitative results of all methods trained on Gaussians of ARKitScenes and tested on MultiScan.}
  \label{fig:arkit2multiscan}
\end{figure*}

\noindent\textbf{Models trained on Gaussians of ARKitScens:} in Figure~\ref{fig:gs_arkit2scans}\&\ref{fig:gs_arkit2arkit}\&\ref{fig:arkit2multiscan}, we show more qualitative results of Group 2.

\end{document}